\documentclass[11pt]{article}

\usepackage[preprint]{acl}

\usepackage{times}
\usepackage{latexsym}

\usepackage[T1]{fontenc}

\usepackage[utf8]{inputenc}

\usepackage{microtype}

\usepackage{inconsolata}

\usepackage{graphicx}
\usepackage{amssymb}
\usepackage{booktabs}
\usepackage{multirow}
\usepackage{amsmath}

\title{GHI: Graphormer over Conditioned Hypergraph Incidence for Aspect-Based Sentiment Analysis}

\author{Yu Du \\
  Qiqihar University \\
  \texttt{2025936317@qqhru.edu.cn} \\\And
  Wenlong Zhu \\
  Qiqihar University \\
  \texttt{zwl\_qqhr@qqhru.edu.cn} \\\And
  Xingze Li \\
  Qiqihar University \\
  \texttt{2025936326@qqhru.edu.cn} \\\AND
  Chenglong Cao \\
  Qiqihar University \\
  \texttt{2025936311@qqhru.edu.cn} \\\And
  Jing Wang \\
  Qiqihar University \\
  \texttt{2025912341@qqhru.edu.cn} \\\And
  Yukun Ma \\
  Qiqihar University \\
  \texttt{2024935328@qqhru.edu.cn}
  \\}

\begin{document}
\maketitle
\begin{abstract}
Aspect-based sentiment analysis (ABSA) requires models to bind sentiment evidence to the correct aspect, making it a natural testbed for fine-grained structural reasoning. We introduce \textsc{GHI}, a Graphormer-over-Conditioned-Hypergraph-Incidence framework that is designed as an incidence-based structural reasoning layer built on a bipartite topology. GHI represents diverse linguistic and semantic evidence as token--hyperedge incidence relations, allowing different structural signals to be incorporated  through a unified interface. Extensive experiments on six standard ABSA benchmarks show that \textsc{GHI} outperforms all baselines on the SemEval domains, and multi-seed evaluations show stable improvements over strong DeBERTa. Further experiments show that with only 247M parameters, \textsc{GHI} approaches the performance of 11B Flan-T5 based methods on the ISE benchmark. Moreover, it demonstrates strong robustness on the challenging ARTS datasets, maintaining highly competitive performance where traditional models degrade. These results demonstrate that compact structural reasoning remains a valuable alternative to scale-driven approaches for fine-grained tasks.
\end{abstract}

\section{Introduction}

\begin{figure}[t]
  \includegraphics[width=\columnwidth]{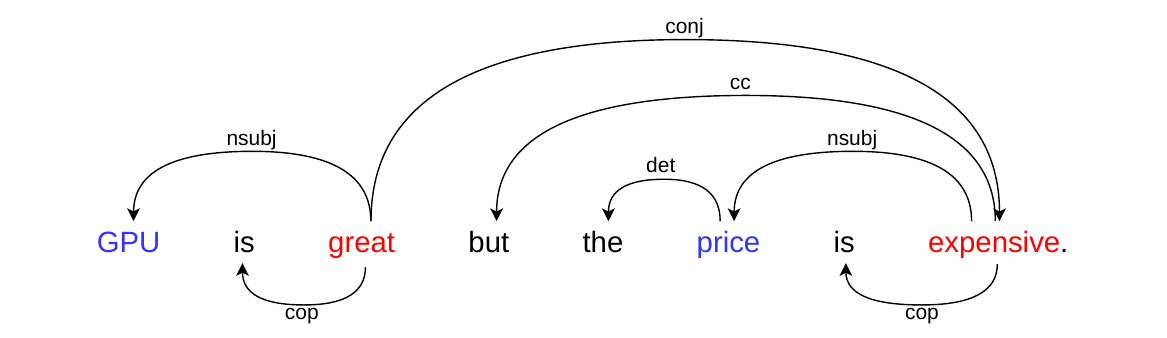}
  \caption{An example sentence with two different aspects (colored in blue) and opinion evidences (colored in red). The sentence has already been preprocessed by its dependency parser.}
  \label{fig:absa}
\end{figure}

Aspect-Based Sentiment Analysis (ABSA) aims to predict the sentiment polarity toward a given aspect term or target entity \citep{9996141}. Unlike sentence-level sentiment classification, ABSA is a fine-grained evidence-binding task that requires the model to separate different opinion clues within the same sentence. For example, as shown in Figure~\ref{fig:absa}, the sentence expresses a positive sentiment "\textit{great}" towards "\textit{GPU}" while expressing a negative sentiment "\textit{expensive}" towards "\textit{price}", requiring the model to bind each opinion cue to correct aspect.

More fundamentally, to further distinguish different evidence-binding patterns, ABSA needs diverse complex structural representations from different directions. In the local-context direction, \citet{app9163389} constrain attention to aspect-centered windows, making the model focus on nearby opinion words. Recent fusion-based models further combine semantic attention with syntactic structure, for example, \citet{JIN2025111654} use Aspect-NA and adaptive hierarchical cross-attention to integrate semantic and dependency-aware features. Another direction focuses mainly on graph structures. Specifically, \citet{Yin_Zhong_2024} couple graph-view message passing with sequence-view Transformer modeling, jointly capturing syntactic connectivity and semantic interactions.

Viewed across existing methods, ABSA methods that introduce high-order relations or inject complex structures can be seen as helping the model identify and organize the aspect-relevant evidence. Moreover, most of them design each source of evidence in isolated views. This observation motivates a general question: can ABSA benefit from a common structural framework through which diverse evidence can be represented, extended, and reasoned over?

\begin{figure}[]
  \includegraphics[width=\columnwidth]{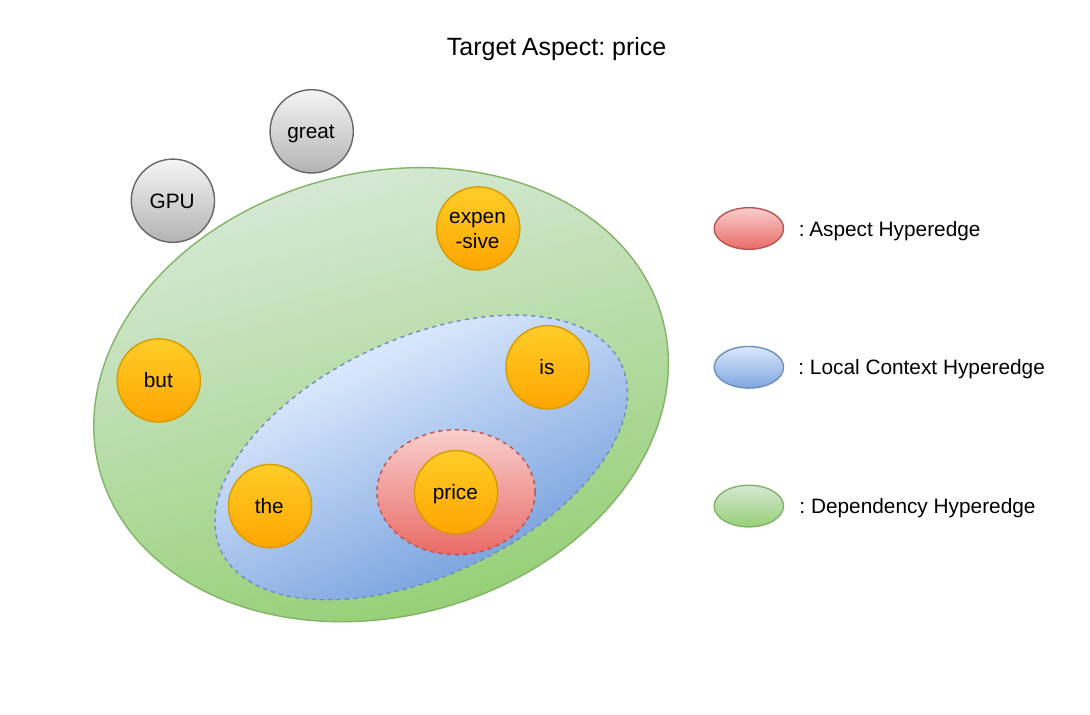}
  \caption{A hypergraph view for the aspect "price". Each colored ellipse denotes a hyperedge that connects multiple tokens as one evidence unit.}
  \label{fig:hypergraph}
\end{figure}

To make this picture clear, we need a structure that possesses  both the capability for multiple representations and scalability. For this reason, hypergraphs provide a natural candidate \citep{Feng_You_Zhang_Ji_Gao_2019}. As illustrated in Figure~\ref{fig:hypergraph}, their hyperedge design enables heterogeneous groups of tokens to be connected under a shared evidence unit, and the structural properties inherited from graphs make them extensible to new sources. Meanwhile, recent studies attempt to construct word-level relational hypergraphs to model high-order relations for ABSA, further showing their potential for high-order reasoning \citep{OUYANG2024123412,JU2025114701,kashyap2025graphshypergraphsenhancingaspectbased}.

However, to model such a complex structure, there remains a lack of effective ways to integrate multi-level information. Another straightforward challenge is that it is typically difficult to modify a graph structure when attempting to incorporate new knowledge. Fortunately, previous works have demonstrated the potential of both global attention and dynamic scalability on complex graph structures. For instance, \citet{NEURIPS2021_f1c15925} proposed Graphormer, which utilizes the shortest path distance (SPD) between nodes to enable global attention for complex graph structure. In parallel, within the computer vision domain, \citet{lei2025yolov13realtimeobjectdetection} introduced the concept of HyperACE. This work demonstrates how adaptive hyperedge mechanisms operate in practice, offering fresh insights into the scalability of hypergraphs.

Building upon these advancements, we propose \textsc{GHI}, a Graphormer-over-Hypergraph-Incidence framework for ABSA. GHI expresses multiple evidence as token--hyperedge incidence relations, so that different structural signals can be incorporated through a unified interface while keeping the downstream reasoning layer unchanged. In our ABSA instantiation, \textsc{GHI} uses a small set of canonical ABSA priors, including aspect spans, aspect-relative local regions, and dependency neighborhoods, and complements them with context-conditioned adaptive hyperedges for sample-specific latent evidence. In addition, by lifting hyperedges into explicit nodes, \textsc{GHI} performs Graphormer-style reasoning over a bipartite star-expanded token--hyperedge graph.

In summary, the main contributions of our work are as follows:
\begin{itemize}
    \item We propose \textsc{GHI}, an incidence-based structural reasoning framework for ABSA. \textsc{GHI} represents linguistic and semantic evidence as token--hyperedge incidence relations, providing a unified interface that can naturally accommodate different structural signals without source-specific reasoning branches.

    \item We introduce a bipartite star-expanded Graphormer built on a static-adaptive hypergraph design. By lifting diverse evidence into explicit reasoning nodes, \textsc{GHI} turns token--hyperedge incidence relations into a bipartite topology and applies Graphormer-style global attention over it.

    \item We conduct comprehensive evaluations on standard ABSA benchmarks, implicit sentiment evaluation (ISE), and adversarial robustness tests (ARTS). Results show that \textsc{GHI} outperforms strong baselines while exhibiting robust performance against complex linguistic variations.
\end{itemize}

\section{Methodology}

\subsection{Overview}

\begin{figure*}[t]
  \centering
  \includegraphics[width=0.75\linewidth]{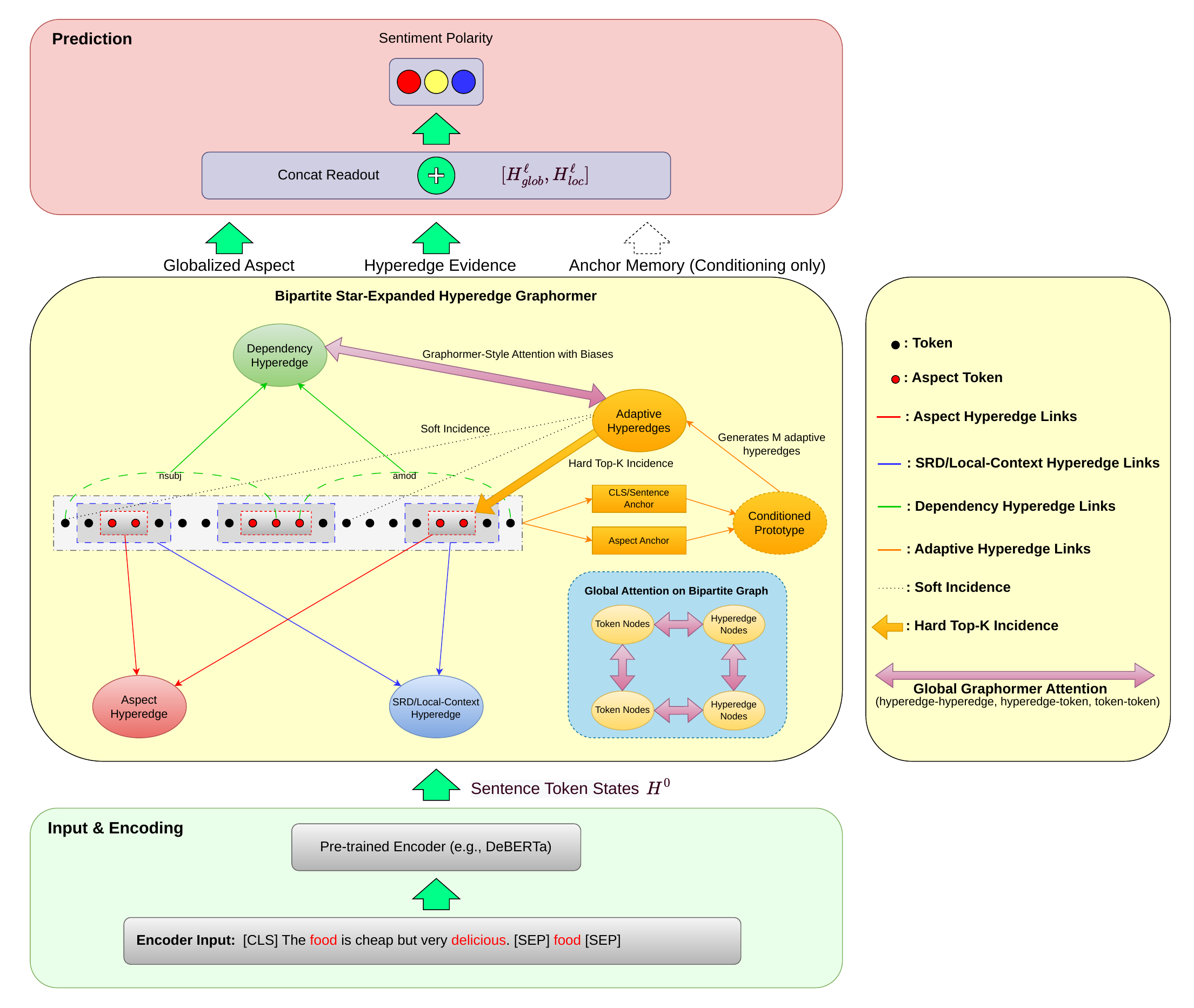}
  \caption {The overall architecture of the proposed \textsc{GHI} framework. Static ABSA priors, including aspect (red), SRD/local-context (blue), and dependency (green) hyperedges, are color-coded by source and represented together with adaptive hyperedges in a token--hyperedge incidence structure. Dotted lines denote soft incidence weights used for token--hyperedge propagation, while orange Top-$K$ links define the sparse hard incidence topology. The purple bidirectional arrows indicate Graphormer-style global attention over the entire bipartite token--hyperedge graph.}
  \label{fig:overall_structure}
\end{figure*}

We propose \textbf{GHI}, a \textbf{G}raphormer-over-Conditioned \textbf{H}ypergraph \textbf{I}ncidence framework for ABSA. \textsc{GHI} lifts structural priors and adaptive semantic clusters into explicit nodes, thereby forming a bipartite star-expanded graph for global Graphormer attention (indicated by the purple arrows). As illustrated in Figure \ref{fig:overall_structure}, the framework operates as a unified incidence-level routing layer, where distinct colors differentiate the diverse evidence sources.

Specifically, \textsc{GHI} constructs static hyperedges for deterministic linguistic evidence and adaptive hyperedges conditioned on contextual anchors. With both soft and hard Top-$K$ incidence views, the soft incidence supports differentiable token--hyperedge propagation, while the hard Top-$K$ incidence instantiates a sparse star-expanded topology for Graphormer reasoning with structural biases.

\subsection{Task Formulation and Encoding}

Given an aspect span $a = [l, r)$ within a sequence $x = [x_{1}, \ldots, x_{N}]$, and a sentiment label $y \in \mathcal{Y}$, ABSA aims to predict the sentiment polarity expressed toward the given aspect. For pre-trained encoders, the sentence--aspect pair is used as the input, while only the contextual states of the original sentence tokens are retained as $H^0$ for graph reasoning. We encode the sentence-aspect pair to obtain contextual representations:

\begin{equation}
  \label{eq:1}
  H^{0} = [h_{1}^{0}, \ldots, h_{N}^{0}] = \mathrm{Enc}(x, a),
\end{equation}

where $H^{0} \in \mathbb{R}^{N \times d}$, and the half-open span $[l,r)$ denotes the target aspect tokens from $x_l$ to $x_{r-1}$.

Graph reasoning operates solely on valid sentence tokens. With a binary mask $m \in \{0, 1\}^{N}$, the sentence anchor is initialized as $c^{0} = \mathrm{Pool}_{m}(H^{0})$ with mean pooling $\mathrm{Pool}_m(\cdot)$ over valid tokens, and the aspect anchor is initialized as $a^{0} = \mathrm{Pool}_{[l, r)}(H^{0})$ with mean pooling $\mathrm{Pool}_{[l,r)}(\cdot)$ over the target aspect span.

\label{sec:task}

\subsection{Conditioned Hypergraph Incidence}

The core of \textsc{GHI} is a conditioned hypergraph incidence representation. At layer $\ell$, we define a hypergraph $\mathcal{G}^{\ell} = (V, \mathcal{E}^{\ell}, I^{\ell})$ over the current token states $H^{\ell}$, where $V$ is the set of token nodes, $\mathcal{E}^{\ell}$ is the set of hyperedges, and $I^{\ell} \in \mathbb{R}^{\lvert V \rvert \times \lvert \mathcal{E}^{\ell} \rvert}$ is the token--hyperedge incidence matrix. \textsc{GHI} combines task-informed static hyperedges with context-conditioned adaptive hyperedges. The sentence and aspect anchors are maintained as layer-wise memories, after each reasoning layer $\ell$, they are updated by gated MLPs as $c^{\ell+1}, a^{\ell+1} = \mathrm{AnchorUpdate}(c^\ell,a^\ell,H^{\ell+1})$, where $\mathrm{AnchorUpdate}(\cdot)$ denotes gated MLP updates over the aspect-level pooled regions.

\paragraph{Static Hyperedges}

We first construct a binary incidence matrix $I^{\mathrm{sta}}$ from three static hyperedge priors. The aspect hyperedge $e_{\mathrm{asp}}$ connects all tokens inside the target span: $e_{\mathrm{asp}} = \{i \mid l \le i < r\}$. To encode aspect-centered local context, we follow \citet{app9163389} and compute the semantic-relative distance (SRD) $d_i = \mathrm{max}\big(0, \big \lvert i - \frac{l + r - 1}{2} \big \rvert - \big \lfloor \frac{r - l}{2} \big \rfloor \big)$. This prior is then instantiated as a local-context hyperedge $e_{\mathrm{srd}}$ that connects tokens within a radius $\rho$: $e_{\mathrm{srd}} = \{i \mid m_{i} = 1, d_{i} \le \rho\}$. Then, the dependency hyperedge $e_{\mathrm{dep}}$ collects tokens reachable from aspect tokens within $T$ hops on the dependency graph: $e_{\mathrm{dep}} = \{i \mid m_{i} = 1, \mathrm{dist}_{\mathrm{dep}}(i, e_{\mathrm{asp}}) \le T\}$. Word-level dependency edges are projected to subwords to align with encoder outputs. Taken together, these priors form a static incidence matrix $I^{\mathrm{sta}}\in\mathbb{R}^{|V|\times S}$, where $S$ denotes the number of static hyperedges.

\paragraph{Adaptive Hyperedges}
\label{sec:adaptive_hyperedges}

Static hyperedges provide reliable task priors, but they cannot cover all sample-specific opinion patterns. To complement them, \textsc{GHI} induces a small set of adaptive hyperedges at each layer, conditioned on the locally refined token states and current anchor memories $c^\ell$ and $a^\ell$. As illustrated in the upper-right part of Figure~\ref{fig:overall_structure}, the adaptive hyperedge prototypes generate $M$ adaptive hyperedges, and further form a soft token--hyperedge incidence matrix $I^{\ell}_{\mathrm{ad}} \in \mathbb{R}^{\lvert V \rvert \times M}$:

\begin{equation}
  \label{eq:2}
  I^{\ell}_{\mathrm{ad}} = \mathrm{AdaptiveIncidence}(\widetilde{H}^{\ell}, c^{\ell}, a^{\ell}),
\end{equation}

where $\mathrm{AdaptiveIncidence}(\cdot)$ denotes the adaptive incidence generator. Its prototype-based parameterization is given in Appendix~\ref{appendix:adaptive_hyperedges_generating}.

The adaptive incidence is concatenated with static incidence matrix to support differentiable token--hyperedge propagation, generally forming a soft incidence matrix $I_{\mathrm{soft}}^{\ell} \in \mathbb{R}^{\lvert V \rvert \times (S+M)} = [I^{\mathrm{sta}}, I_{\mathrm{ad}}^{\ell}]$. Meanwhile, we retain the Top-$K$ tokens for each adaptive hyperedge to obtain a sparse hard incidence matrix $I_{\mathrm{hard}}^{\ell} \in \mathbb{R}^{\lvert V \rvert \times (S+M)} = [I^{\mathrm{sta}}, \mathrm{TopK}({I}_{\mathrm{ad}}^{\ell})]$, which instantiates the bipartite star-expanded topology used by Graphormer reasoning.

\subsection{GHI Reasoning Layer}
\label{sec:local_context_refinement}

GHI stacks $L$ reasoning layers over the conditioned incidence structure. A layer-wise computation flow is provided in Appendix~\ref{appendix:layer}. By treating both tokens and hyperedges as explicit reasoning nodes, each layer couples two complementary views of token--hyperedge reasoning: the soft incidence view supports differentiable propagation over graded token--hyperedge memberships, while the hard incidence view instantiates a sparse token--hyperedge bipartite topology. This star-expanded graph allows Graphormer-style attention to model token--token, token--hyperedge, and hyperedge--hyperedge interactions within a shared structural space.

\paragraph{Local Context Refinement}

Before constructing adaptive incidence, \textsc{GHI} applies a local-window self-attention to refine short-range token interactions. Given the graph-visible mask $m$ and window size $w$, $\mathrm{LocalAttn}(\cdot)$ restricts multi-head self-attention to token pairs $(i,j)$ satisfying $m_i=m_j=1$ and $\lvert i-j \rvert \leq w$. The local refinement is written as: $\widetilde{H}^{\ell} = H^{\ell} + \mathrm{LocalAttn(\mathrm{LN}}(H^{\ell}), m, w)$. The locally refined states $\widetilde{H}^{\ell}$ then guide the adaptive hyperedge incidence described in Section~\ref{sec:adaptive_hyperedges}.

\paragraph{Incidence-Aware Hypergraph Reasoning}

Given the soft incidence matrix $I_{\mathrm{soft}}^{\ell}$, GHI summarizes token information into hyperedge states through incidence-weighted pooling:

\begin{equation}
  \label{eq:3}
  Z^{\ell} = \mathrm{EdgePool}(\widetilde{H}^{\ell}, I_{\mathrm{soft}}^{\ell}).
\end{equation}

Here, $\mathrm{EdgePool}(\cdot)$ aggregates token states according to their soft token--hyperedge participation weights and adaptive priors.

We denote the resulting incidence-aware local token representation as $H^{\ell}_{\mathrm{loc}}$. It is obtained by two complementary operations. $\mathrm{HGRefine}(\cdot)$ first performs soft token--hyperedge propagation and writes hyperedge messages back to tokens, while Relation-Aware Incidence Attention $\mathrm{IncAttn}(\cdot)$ applies  token--hyperedge attention using incidence-level relation features $\Phi^{\ell}$:

\begin{equation}
  \label{eq:4}
  \begin{aligned}
  H_{\mathrm{loc}}^{\ell} = \widetilde{H}^{\ell} + \mathrm{HGRefine}(\widetilde{H}^{\ell}, Z^{\ell}, I_{\mathrm{soft}}^{\ell}) \\
  + \mathrm{IncAttn}(\widetilde{H}^{\ell}, Z^{\ell}, I_{\mathrm{soft}}^{\ell}, \Phi^{\ell})
  \end{aligned}
\end{equation}

Specifically, $\Phi^{\ell}$ includes edge type, edge role (static or adaptive), incidence weight, and SRD.

\paragraph{Star-Expanded Graphormer Reasoning}

While the soft incidence view supports differentiable hypergraph propagation, \textsc{GHI} also uses the hard incidence view $I_{\mathrm{hard}}^{\ell}$ to construct a sparse star-expanded graph. The expanded graph contains both token nodes and hyperedge nodes. A token node is connected to a hyperedge node if the corresponding hard incidence entry is non-zero. The node states of the star-expanded graph are initialized as $X^{\ell} = [\widetilde{H}^{\ell}; Z^{\ell}]$.

Following the Graphormer design \citep{NEURIPS2021_f1c15925}, we extract structural encodings from the expanded graph, including topological connectivity and pairwise relation types, and inject them as Graphormer structural biases for multi-head self-attention. In this way, global attention is aware of token--token, token--hyperedge, and hyperedge--hyperedge relations within the bipartite topology. The global update is written compactly with layer normalization (LN):

\begin{equation}
  \label{eq:5}
  X_{\mathrm{glob}}^{\ell} = X^{\ell} + \mathrm{GraphormerAttn}(\mathrm{LN}(X^{\ell}), \mathcal{B}^{\ell}),
\end{equation}

where $\mathcal{B}^{\ell}$ denotes the Graphormer structural biases derived from the hard incidence topology $I_{\mathrm{hard}}^{\ell}$. We then split the output back into token and hyperedge parts:

\begin{equation}
  \label{eq:6}
  X_{\mathrm{glob}}^{\ell} = [H_{\mathrm{glob}}^{\ell}; Z^{\ell + 1}],
\end{equation}

where \(H_{\mathrm{glob}}^\ell\) denotes the global token representation produced by the star-expanded Graphormer.

\paragraph{Local-Global Fusion}

\textsc{GHI} fuses the local evidence ${H}_{\mathrm{loc}}^{\ell}$ and the star-expanded global token representation conditioned on the sentence and aspect anchors:

\begin{equation}
  \label{eq:7}
  U^{\ell} = \mathrm{Fuse}(H_{\mathrm{glob}}^{\ell}, {H}_{\mathrm{loc}}^{\ell}, c^{\ell}, a^{\ell}).
\end{equation}

The next-layer token states are then produced by a residual feed-forward update: $H^{\ell + 1} = U^{\ell} + \mathrm{FFN}(\mathrm{LN}(U^{\ell}))$.

\subsection{Prediction and Training}

After $L$ reasoning layers, \textsc{GHI} predicts sentiment polarity from local readout derived from $H^{\ell}_{loc}$ and global readout derived from $H^{\ell}_{glob}$. Two readouts are concatenated and passed to a linear classifier to obtain $p(y \mid x, a)$. Notably, in our main configuration, the anchor memory is used for conditioning incidence induction and local-global fusion $U^\ell$, but is not included in the final readout. The entire training is with the standard cross-entropy loss:

\begin{equation}
  \label{eq:8}
  \mathcal{L} = -\log p(y \mid x,a),
\end{equation}

\section{Experiments}

We conduct experiments under a unified evaluation protocol. Unless otherwise specified, we use the standard train / test splits and report Accuracy and Macro-F1. We also conduct hyperparameter sensitivity analyses in Appendix~\ref{appendix:sensitivity}. All experimental details are provided in Appendix~\ref{appendix:experimental_details}.

\subsection{Main Results}

\begin{table*}
  \centering
  \resizebox{\textwidth}{!}{
  \begin{tabular}{ll *{12}{c}}
    \hline
    \multirow{2}{*}{Model} & \multirow{2}{*}{Encoder} 
    & \multicolumn{2}{c}{Restaurant14} & \multicolumn{2}{c}{Laptop} 
    & \multicolumn{2}{c}{Twitter} & \multicolumn{2}{c}{Restaurant15} 
    & \multicolumn{2}{c}{Restaurant16} & \multicolumn{2}{c}{MAMS} \\

    \cmidrule(lr){3-4} \cmidrule(lr){5-6} \cmidrule(lr){7-8} \cmidrule(lr){9-10} \cmidrule(lr){11-12} \cmidrule(lr){13-14}
    
    & & Acc & F1 & Acc & F1 & Acc & F1 & Acc & F1 & Acc & F1 & Acc & F1 \\
    \hline
    SK-GCN\citep{ZHOU2020106292} & BERT & 83.48 & 75.19 & 79.00 & 75.57 & 75.00 & 73.01 & 83.20 & 66.78 & 87.19 & 72.02 & - & - \\
    KumaGCN\citep{chen-etal-2020-inducing} & BERT & 86.43 & 80.30 & 81.98 & 78.81 & 77.89 & 77.03 & 86.35 & 70.76 & 92.53 & 79.24 & - & - \\
    R-GAT\citep{wang-etal-2020-relational} & BERT & 86.60 & 81.35 & 78.21 & 74.07 & 76.15 & 74.88 & - & - & - & - & 84.52 & 83.74 \\
    DGEDT\citep{tang-etal-2020-dependency} & BERT & 86.30 & 80.00 & 79.80 & 75.60 & 77.90 & 75.40 & 84.00 & 71.00 & 91.90 & 79.00 & - & - \\
    DualGCN\citep{li-etal-2021-dual-graph} & BERT & 87.13 & 81.16 & 81.80 & 78.10 & 77.40 & 76.02 & - & - & - & - & - & - \\
    TGCN\citep{tian-etal-2021-aspect} & BERT & 86.16 & 79.95 & 80.88 & 77.03 & 76.45 & 75.25 & 85.26 & 71.69 & 92.32 & 77.29 & 83.38 & 82.77 \\
    SSEGCN\citep{zhang-etal-2022-ssegcn} & BERT & 87.31 & 81.09 & 81.01 & 77.96 & 77.40 & 76.02 & - & - & - & - & - & - \\
    dotGCN\citep{chen-etal-2022-discrete} & BERT & 86.16 & 80.49 & 81.03 & 78.10 & 78.11 & 77.00 & 85.24 & 72.74 & 93.18 & \underline{82.32} & 84.95 & 84.44 \\
    MWGCN\citep{Yu2023MWGCN} & BERT & 86.36 & 80.54 & 79.78 & 76.68 & 75.00 & 74.30 & 85.61 & 72.88 & 82.05 & 79.21 & - & - \\
    APARN\citep{ma-etal-2023-amr} & BERT & 87.76 & 82.44 & 81.96 & 79.10 & \textbf{79.76} & \textbf{78.79} & - & - & - & - & \textbf{85.59} & \textbf{85.06} \\
    DAGCN\citep{wang-etal-2024-dagcn} & BERT & 88.03 & 82.64 & 82.59 & 79.40 & 78.73 & 78.01 & - & - & - & - & 85.25 & 84.87 \\
    $\mathrm{S^{2}GSL}$\citep{chen-etal-2024-s2gsl} & BERT & 87.31 & 82.84 & 82.46 & 79.07 & 77.84 & 77.11 & - & - & - & - & 85.17 & 84.74 \\
    TextGT\citep{Yin_Zhong_2024} & BERT & 87.31 & 82.27 & 81.33 & 78.71 & 77.70 & 76.45 & - & - & 92.21 & 81.48 & - & - \\
    MLFM\citep{JIN2025111654} & BERT & 87.31 & 82.50 & 82.12 & 79.06 & 77.55 & 76.83 & - & - & - & - & - & - \\
    PWCN$^{\dagger}$\citep{10.1145/3331184.3331351} & RoBERTa & 87.35 & 80.85 & 84.01 & 81.08 & 77.02 & 75.52 & - & - & - & - & - & - \\
    SARL\citep{wang-etal-2021-eliminating-sentiment} & RoBERTa & 88.21 & 82.44 & \underline{85.42} & \underline{82.97} & 78.03 & 76.97 & \underline{88.19} & \underline{73.83} & \underline{94.62} & 81.92 & - & - \\
    PConv\citep{Feng2023PConv} & RoBERTa & 89.29 & 84.27 & 83.54 & 80.89 & 78.47 & 77.53 & - & - & - & - & \underline{85.55} & \underline{85.05} \\
    DAGF\citep{JI2026115040} & RoBERTa & 89.37 & 84.37 & 83.07 & 80.38 & 77.25 & 76.55 & - & - & - & - & - & - \\
    RCL\citep{JIAN2024103539} & DeBERTa & 89.38 & 84.68 & 82.76 & 80.28 & 78.32 & 77.47 & - & - & - & - & - & - \\
    AGCL$^{\ddagger}$\citep{jian-etal-2025-agcl} & DeBERTa & \underline{90.30} & \underline{85.63} & 84.54 & 82.15 & \underline{78.85} & \underline{78.15} & - & - & - & - & - & - \\
    GHI(ours) & DeBERTa & \textbf{90.97} & \textbf{86.40} & \textbf{86.08} & \textbf{83.74} & 78.73 & 77.72 & \textbf{90.22} & \textbf{79.37} & \textbf{95.29} & \textbf{87.84} & 84.96 & 84.42 \\
    \hline
  \end{tabular}
  }
  \caption{Overall performance on six ABSA benchmark datasets. Best results are in bold and second-best are underlined. $\dagger$ indicates result retrieved from \citet{dai-etal-2021-syntax}, while others are from their original papers. $\ddagger$ denotes that AGCL \citep{jian-etal-2025-agcl} utilizes a frozen DeBERTa-Large encoder rather than the base version.}
  \label{tab:main_results}
\end{table*}

Table~\ref{tab:main_results} reports the overall performance of \textsc{GHI} alongside previous state-of-the-art models across six benchmark datasets. \textsc{GHI} particularly outperforms all baselines on the SemEval domains. Driven by the DeBERTa encoder \citep{DBLP:journals/corr/abs-2006-03654}, the proposed framework yields 90.97\% / 86.40\% of Accuracy / Macro-F1 on Restaurant14 and 86.08\% / 83.74\% on Laptop. Notably, \textsc{GHI} outperforms knowledge-augmented baselines such as AGCL \citep{jian-etal-2025-agcl}, suggesting that incidence-based structural modeling can also provide gains complementary to external augmentation. Furthermore, \textsc{GHI} maintains competitive performance on multi-domain datasets, including Twitter and the challenging MAMS benchmark. Overall, \textsc{GHI} shows strong gains on the SemEval domains and remains comparable to strong baselines on Twitter and MAMS.

\subsection{Controlled Encoder Comparison}

\begin{table*}[t]
    \centering
    \resizebox{\textwidth}{!}{
    \begin{tabular}{ll *{4}{c}}
    \hline
    \multirow{2}{*}{Encoder} & \multirow{2}{*}{Model} & \multicolumn{2}{c}{Restaurant14} & \multicolumn{2}{c}{Laptop}  \\
     
    \cmidrule(lr){3-4} \cmidrule(lr){5-6}

     & & Acc & F1 & Acc & F1 \\
    \hline
    \multirow{5}{*}{$\mathrm{BERT}_{\mathrm{base}}$} & BERT & 85.65(0.61) & 79.13(1.21) & 78.39(0.67) & 74.72(0.73) \\
     & DAGCN & $86.17(0.42)_{\uparrow 0.52}$ & $79.59(0.80)_{\uparrow 0.46}$ & $80.36(0.46)_{\uparrow 1.97}$ & $76.90(0.67)_{\uparrow 2.18}$ \\
     & TextGT & $86.38(0.44)_{\uparrow 0.73}$ & $80.39(0.70)_{\uparrow 1.26}$ & $80.70(0.64)_{\uparrow 2.31}$ & $77.65(0.86)_{\uparrow 2.93}$ \\
     & MLFM & $86.23(0.48)_{\uparrow 0.58}$ & $79.93(0.71)_{\uparrow 0.80}$ & $80.47(0.56)_{\uparrow 2.08}$ & $76.92(0.90)_{\uparrow 2.20}$ \\
     & GHI(ours) & $86.06(0.99)_{\uparrow 0.41}$ & $80.30(1.13)_{\uparrow 1.15}$ & $79.59(0.79)_{\uparrow 1.20}$ & $76.11(0.90)_{\uparrow 1.39}$ \\
    \hline
    \multirow{5}{*}{$\mathrm{DeBERTa}_{\mathrm{base}}$} & DeBERTa & 89.33(1.05) & 83.74(1.64) & 83.67(0.45) & 80.78(0.41) \\
     & DAGCN & $89.49(0.30)_{\uparrow 0.16}$ & $84.25(0.42)_{\uparrow 0.51}$ & $84.46(0.65)_{\uparrow 0.79}$ & $81.66(0.50)_{\uparrow 0.88}$ \\
     & TextGT & $89.97(0.31)_{\uparrow 0.64}$ & $85.30(0.38)_{\uparrow 1.56}$ & $84.84(0.51)_{\uparrow 1.17}$ & $82.33(0.66)_{\uparrow 1.55}$ \\
     & MLFM & $89.63(0.61)_{\uparrow 0.30}$ & $84.62(0.79)_{\uparrow 0.88}$ & $84.59(1.10)_{\uparrow 0.92}$ & $81.73(1.22)_{\uparrow 0.95}$ \\
     & GHI(ours) & $\textbf{90.33(0.49)}_{\uparrow 1.00}$ & $\textbf{85.65(0.77)}_{\uparrow 1.91}$ & $\textbf{85.06(0.70)}_{\uparrow 1.39}$ & $\textbf{82.63(0.71)}_{\uparrow 1.85}$ \\
    \hline
    \end{tabular}
    }
    \caption{Controlled comparison under unified settings on SemEval-2014 domains. All results are reported as mean (standard deviation) over 5 seeds. Models above are all implemented by us. 
}
    \label{tab:reproduction}
\end{table*}

Previous studies have highlighted reproducibility issues and evaluation instability in ABSA evaluation, showing that random seeds, encoder capacities, and implementation details can noticeably affect reported performance \citep{dai-etal-2021-syntax,10.1007/978-3-030-72240-1_7,10.1145/3583780.3614752}. To address these concerns, we evaluate \textsc{GHI} alongside representative structural models under identical BERT-base and DeBERTa-base encoders across multiple random seeds, aiming to separate structural gains from encoder capacity and training variance.

Table~\ref{tab:reproduction} presents the controlled evaluation results. We strictly adhere to the hyperparameter configurations recommended in the original papers, and use the multi-seed setting to observe relative improvements under the same encoder constraints. Under these conditions, \textsc{GHI} yields stable and substantial gains. Specifically, it improves Accuracy / Macro-F1 by 1.00\% / 1.91\% on Restaurant14 and 1.39\% / 1.85\% on Laptop relative to the strong vanilla DeBERTa baseline. Furthermore, \textsc{GHI} maintains competitive performance when compared to other state-of-the-art models under the same encoder constraints. These results verify that the gains of \textsc{GHI} are not merely attributable to encoder capacity or training variance, but are consistent with the benefits of its incidence-based reasoning design.

\subsection{Ablations}

\begin{table*}[t]
    \centering
    \resizebox{\textwidth}{!}{
    \begin{tabular}{l *{6}{c}}
    \hline
    \multirow{2}{*}{Model} & \multicolumn{2}{c}{Restaurant14} & \multicolumn{2}{c}{Laptop} & \multicolumn{2}{c}{Twitter} \\
    \cmidrule(lr){2-3} \cmidrule(lr){4-5} \cmidrule(lr){6-7}
     & Acc & F1 & Acc & F1 & Acc & F1 \\
    \hline
    GHI & \textbf{90.97} & \textbf{86.40} & \textbf{86.08} & \textbf{83.74} & \textbf{78.73} & \textbf{77.72} \\
    w/o Adaptive Hyperedges & 90.08 & 85.25 & 84.49 & 81.60 & 75.63 & 74.81 \\
    w/o Hyperedge Nodes & 89.45 & 84.14 & 84.18 & 80.39 & 75.63 & 74.99 \\
    w/o Incidence Attention & 88.29 & 83.72 & 84.02 & 80.94 & 76.96 & 76.05 \\
    w/o Graphormer Structural Bias & 89.37 & 83.87 & 84.18 & 81.58 & 76.22 & 75.95 \\
    \hline
    \end{tabular}
    }
    \caption{Ablation results. The variants remove adaptive hyperedges, explicit hyperedge nodes, Relation-Aware Incidence Attention in Eq.~\ref{eq:4}, or Graphormer structural biases $\mathcal{B}^{\ell}$ in Eq.~\ref{eq:5}, respectively.}
    \label{tab:ablations}
\end{table*}

Table~\ref{tab:ablations} presents the ablation results of \textsc{GHI}. On the SemEval-2014 domains, w/o Incidence Attention or w/o Hyperedge Nodes yield the most severe drops. Specifically, w/o Incidence Attention causes the largest accuracy decrease on both datasets, and reduces Macro-F1 by 2.68\% on Restaurant14. In parallel, w/o Hyperedge Nodes proves most harmful to Laptop, reducing its Macro-F1 by 3.35\%. These results suggest that token--hyperedge relations should not be treated merely as transient aggregation paths. Incidence Attention provides relation-aware token--hyperedge routing, allowing the model to distinguish deterministic syntactic links from softer semantic memberships. Meanwhile, the drop caused by removing Hyperedge Nodes supports the motivation of lifting structural priors into explicit reasoning nodes, where token--hyperedge and hyperedge--hyperedge interactions can be modeled directly.

The degradation pattern shifts significantly on the Twitter dataset. Unlike formal SemEval domains, Twitter texts are typically shorter, highly informal, and syntactically noisy, which severely diminishes the reliability of parser-derived fixed dependency priors. In this setting, removing Adaptive Hyperedges incurs the largest performance penalty. This suggests that sample-specific semantic clusters, dynamically induced from token and anchor states, become the decisive factor when explicit syntactic regularities are weak.

\subsection{Implicit Sentiment Evaluation}

\begin{table}[]
    \centering
    \resizebox{\columnwidth}{!}{
    \begin{tabular}{lcc}
    \hline
    Method & Restaurant14 & Laptop \\
    \hline
    BERTAsp+SCAPT\citep{li-etal-2021-learning-implicit} & 72.28 & 77.59 \\
    DeBERTa+SPC$^{\spadesuit}$ & 73.72 & 77.26 \\
    ABSA-ESA\citep{Ouyang_Yang_Liang_Wang_Wang_Li_2024} & 73.76 & 77.91 \\
    GPT3+THOR\citep{fei-etal-2023-reasoning} & 76.55 & 73.12 \\
    Flan-T5+Prompt(11B)\citep{fei-etal-2023-reasoning} & 75.10 & 78.91 \\
    Flan-T5+THOR(11B)\citep{fei-etal-2023-reasoning} & \textbf{79.73} & \textbf{82.43} \\
    DeBERTa+GHI(ours, 247M) & \underline{79.64} & \underline{81.96} \\
    \hline
    \end{tabular}
    }
    \caption{ISE Macro-F1 results. $\spadesuit$ denotes baseline results implemented by us.}
    \label{tab:ise}
\end{table}

Table~\ref{tab:ise} evaluates \textsc{GHI} on the Implicit Sentiment Eval (ISE) benchmark \citep{li-etal-2021-learning-implicit}, a notoriously difficult setting where sentences lack explicit opinion words (e.g., "\textit{The battery lasts only 2 hours.}"). With only 247M parameters, the DeBERTa-based \textsc{GHI} yields an F1 score of 79.64\% on Restaurant14 and 81.96\% on Laptop, closely approaching the 11B-parameter Flan-T5+THOR \citep{fei-etal-2023-reasoning}, a method specifically prompted for multi-step chain-of-thought reasoning. Furthermore, \textsc{GHI} outperforms other heavy-weight baselines, including GPT3+THOR and standard Flan-T5 prompting. These results suggest that the proposed incidence-based reasoning provides competitive implicit-sentiment modeling without relying on large prompted language models.

\subsection{Aspect Robustness Test}

\begin{table}[]
    \centering
    \resizebox{\columnwidth}{!}{
    \begin{tabular}{lcccc}
    \hline
    \multirow{2}{*}{Method} & \multicolumn{2}{c}{Restaurants14-ARTS} & \multicolumn{2}{c}{Laptop14-ARTS} \\
    \cmidrule(lr){2-3} \cmidrule(lr){4-5}
     & Acc & F1 & Acc & F1 \\
    \hline
    DeBERTa$^{*}$ & 74.97 & 66.48 & 67.71 & 65.60 \\
    \hline
    CEIB\citep{Chang_Yang_Jiang_Xu_2024} & 80.00 & 73.97 & 69.18 & 65.51 \\
    LSAT\citep{yang-li-2024-modeling} & 80.31 & 71.37 & 73.58 & 69.28 \\
    LSAP\citep{yang-li-2024-modeling} & 81.19 & 72.54 & 73.34 & 68.46 \\
    LSAE\citep{yang-li-2024-modeling} & \textbf{81.55} & \underline{72.95} & \underline{74.47} & \underline{69.79} \\
    GHI(ours) & \underline{81.08} & \textbf{76.27} & \textbf{76.13} & \textbf{74.12} \\
    \hline
    \end{tabular}
    }
    \caption{Aspect Robustness Test results. Scores by model with $*$ are copied from \citet{yang-li-2024-modeling}.}
    \label{tab:arts}
\end{table}

To evaluate the robustness of \textsc{GHI} in the face of textual adversarial attacks, we employ existing adversarial attack datasets, specifically Laptop14-ARTS and Restaurant14-ARTS \citep{xing-etal-2020-tasty}. Table~\ref{tab:arts} presents results on the ARTS benchmark. \textsc{GHI} achieves the best Macro-F1 on two datasets, improving over strong LSAE \citep{yang-li-2024-modeling} by 3.32\% and 4.33\%, respectively, underscoring the robustness of \textsc{GHI} under challenging settings.

\subsection{Analysis and Visualization}

\begin{figure}[!]
  \includegraphics[width=\columnwidth]{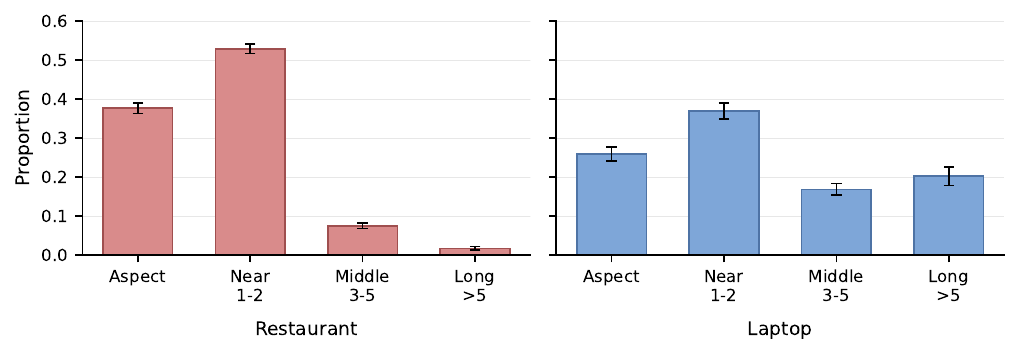}
  \caption{Distance distribution of adaptive Top-$K$ tokens relative to aspect spans on SemEval-14 domains. Error bars denote 95\% bootstrap confidence intervals.}
  \label{fig:adaptive_distance}
\end{figure}

Figure~\ref{fig:adaptive_distance} summarizes the relative positions of tokens retained by adaptive Top-$K$ incidence. The four bins denote tokens inside the aspect span, within 1--2 tokens, 3--5 tokens, and beyond 5 tokens, respectively. The retained tokens are mainly concentrated around the aspect span and its near context, while Laptop keeps more middle and long range tokens than Restaurant14. This result indicates that adaptive hyperedges select different evidence ranges across datasets under the same adaptive topology.

\begin{figure}[!]
  \includegraphics[width=\columnwidth]{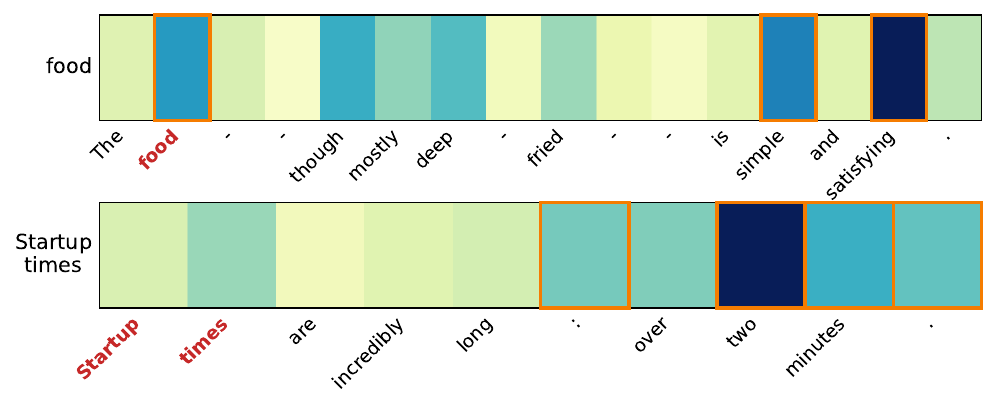}
  \caption{Visualization examples in two cases}
  \label{fig:heat}
\end{figure}

Figure~\ref{fig:heat} presents two illustrative cases to show how \textsc{GHI} organizes aspect-relevant evidence. Orange boxes indicate the hard Top-$K$ tokens retained for the sparse adaptive topology. In both examples, GHI assigns higher weights to informative evidence regions. Notably, in the first case, the adaptive incidence for "\textit{food}" assigns high weights to distant opinion evidence such as "\textit{simple}" and "\textit{satisfying}", suggesting that adaptive hyperedges can group aspect-relevant evidence across a broader context through learned incidence patterns. Furthermore, in the second case, for aspect "\textit{Startup times}", \textsc{GHI} does not rely on the explicit modifier "\textit{long}", but assigns stronger weights to the concrete temporal evidence "\textit{two minutes}". This indicates that the learned incidence can capture multi-token and even implicit evidence expressions toward the target aspect.

\section{Related Work}

Structural modeling remains important for ABSA, where models must bind sentiment evidence to the correct aspect. Recent methods refine syntactic, semantic, and aspect-specific structures from different perspectives.

Early efforts in this direction, such as \citet{app9163389}, pioneered local attention approaches via LCF mechanism. Building upon this, recognizing that sentiment reasoning cannot only rely on a narrow local window, \citet{JIN2025111654} expands this mechanism by fusing semantic and syntactic information through aspect-centered and hierarchical attentions.

Moving beyond optimizations restricted to purely sequential text methods like localized windows, graph-based modeling has emerged as a mainstream paradigm for structural representation. To effectively integrate local mechanisms into graph structures, \citet{wang-etal-2024-dagcn} filters noisy dependency edges with Distance-based Syntactic Weight and Aspect-Fusion Attention. To bridge these graphical structures with powerful global attention, \citet{Yin_Zhong_2024} couples a GNN-based graph view with a Transformer-based sequence view.

While traditional graphs effectively capture pairwise syntax, hypergraph-based ABSA further explores high-order relations beyond pairwise dependency edges. \citet{OUYANG2024123412} builds word-level relational hypergraphs from syntactic and semantic relations and applies aspect-specific hypergraph attention, while \citet{kashyap2025graphshypergraphsenhancingaspectbased} induces dynamic aspect-opinion hyperedges through sample-specific hierarchical clustering. These works suggest that hyperedges are suitable for representing multi-token sentiment evidence.

Expanding the scope from ABSA-specific architectures, graph representation learning has continued to evolve beyond localized message passing. Graphormer shows that structural encodings can be injected into Transformer attention, allowing global graph-aware interaction beyond local message passing \citep{NEURIPS2021_f1c15925}. Once hyperedges are lifted into nodes, token--token, token--hyperedge, and hyperedge--hyperedge relations can be modeled in the same attention space. Recent adaptive hypergraph designs outside NLP, such as HyperACE in YOLOv13, also highlight the potential of dynamically induced high-order correlations \citep{lei2025yolov13realtimeobjectdetection}.

To rigorously test these diverse modeling approaches, recent studies have introduced more challenging evaluation settings for sentiment reasoning. \citet{fei-etal-2023-reasoning} study implicit sentiment through multi-hop reasoning over implicit aspects, opinions, and polarities. \citet{yang-li-2024-modeling} evaluate robustness on ARTS \citep{xing-etal-2020-tasty}, where distracting sentiment words and aspect-opinion mismatches expose reliance on global sentiment shortcuts. These benchmarks test whether models can reliably bind evidence under implicit, noisy, or adversarial conditions, and are therefore useful for evaluating the reliability of incidence-based structural reasoning.

\section{Conclusion}

In this paper, we proposed \textsc{GHI}, a Graphormer-over-conditioned-Hypergraph-Incidence framework for ABSA. \textsc{GHI} converts heterogeneous linguistic and semantic evidence into token--hyperedge incidence relations, avoiding source-specific reasoning branches while retaining explicit structural control. Through soft incidence propagation and hard incidence bipartite star-expanded Graphormer attention, the framework reasons jointly over diverse evidence in a shared space. Results across comprehensive datasets support the effectiveness of this incidence-centered design.

\section*{Limitations}

\textsc{GHI} currently focuses on aspect-term sentiment classification, where the target aspect is given. 
It does not directly address aspect extraction, opinion extraction, or end-to-end aspect--opinion pair discovery. 
Extending the incidence formulation to full ABSA pipelines would require additional decoding mechanisms or supervision for inducing aspects, opinions, and their sentiment relations.

Second, the effectiveness of \textsc{GHI} can still be influenced by the quality of structural priors and the bounded hyperedge budget. 
Dependency-based hyperedges may inherit noise from parser outputs or preprocessing artifacts, while the fixed number of adaptive hyperedges and the Top-\(K\) hard incidence view may miss weak but useful associations in highly complex sentences.

Finally, the current star-expanded Graphormer uses dense attention over the expanded token--hyperedge graph. 
Although the bounded hyperedge budget keeps the overhead controlled for short ABSA texts, applying the same design to long reviews, documents, or dialogue-level sentiment analysis may require sparse attention or hierarchical incidence construction. 

\bibliography{custom}

@ARTICLE{9996141,
    author={Zhang, Wenxuan and Li, Xin and Deng, Yang and Bing, Lidong and Lam, Wai},
    journal={IEEE Transactions on Knowledge and Data Engineering}, 
    title={A Survey on Aspect-Based Sentiment Analysis: Tasks, Methods, and Challenges}, 
    year={2023},
    volume={35},
    number={11},
    pages={11019-11038},
    doi={10.1109/TKDE.2022.3230975}
}

@inproceedings{wang-etal-2024-dagcn,
    title = "{DAGCN}: Distance-based and Aspect-oriented Graph Convolutional Network for Aspect-based Sentiment Analysis",
    author = "Wang, Zhihao  and
      Zhang, Bo  and
      Yang, Ru  and
      Guo, Chang  and
      Li, Maozhen",
    editor = "Duh, Kevin  and
      Gomez, Helena  and
      Bethard, Steven",
    booktitle = "Findings of the Association for Computational Linguistics: NAACL 2024",
    month = jun,
    year = "2024",
    address = "Mexico City, Mexico",
    publisher = "Association for Computational Linguistics",
    url = "https://aclanthology.org/2024.findings-naacl.120/",
    doi = "10.18653/v1/2024.findings-naacl.120",
    pages = "1863--1876"
}

@article{Yin_Zhong_2024, 
    title={TextGT: A Double-View Graph Transformer on Text for Aspect-Based Sentiment Analysis}, 
    volume={38}, 
    url={https://ojs.aaai.org/index.php/AAAI/article/view/29911}, 
    DOI={10.1609/aaai.v38i17.29911}, 
    abstractNote={Aspect-based sentiment analysis (ABSA) is aimed at predicting the sentiment polarities of the aspects included in a sentence instead of the whole sentence itself, and is a fine-grained learning task compared to the conventional text classification. In recent years, on account of the ability to model the connectivity relationships between the words in one sentence, graph neural networks have been more and more popular to handle the natural language processing tasks, and meanwhile many works emerge for the ABSA task. However, most of the works utilizing graph convolution easily incur the over-smoothing problem, while graph Transformer for ABSA has not been explored yet. In addition, although some previous works are dedicated to using both GNN and Transformer to handle text, the methods of tightly combining graph view and sequence view of text is open to research. To address the above issues, we propose a double-view graph Transformer on text (TextGT) for ABSA. In TextGT, the procedure in graph view of text is handled by GNN layers, while Transformer layers deal with the sequence view, and these two processes are tightly coupled, alleviating the over-smoothing problem. Moreover, we propose an algorithm for implementing a kind of densely message passing graph convolution called TextGINConv, to employ edge features in graphs. Extensive experiments demonstrate the effectiveness of our TextGT over the state-of-the-art approaches, and validate the TextGINConv module. The source code is available at https://github.com/shuoyinn/TextGT.}, 
    number={17}, 
    journal={Proceedings of the AAAI Conference on Artificial Intelligence}, 
    author={Yin, Shuo and Zhong, Guoqiang}, 
    year={2024}, 
    month={Mar.}, 
    pages={19404–19412} 
}

@article{OUYANG2024123412,
    title = {Aspect-based sentiment classification with aspect-specific hypergraph attention networks},
    journal = {Expert Systems with Applications},
    volume = {248},
    pages = {123412},
    year = {2024},
    issn = {0957-4174},
    doi = {https://doi.org/10.1016/j.eswa.2024.123412},
    url = {https://www.sciencedirect.com/science/article/pii/S095741742400277X},
    author = {Jihong Ouyang and Chang Xuan and Bing Wang and Zhiyao Yang}
}

@article{JU2025114701,
    title = {Dual contrastive learning-based hypergraph convolutional network for aspect-based sentiment classification},
    journal = {Knowledge-Based Systems},
    volume = {330},
    pages = {114701},
    year = {2025},
    issn = {0950-7051},
    doi = {https://doi.org/10.1016/j.knosys.2025.114701},
    url = {https://www.sciencedirect.com/science/article/pii/S095070512501740X},
    author = {Xinyi Ju and Ling Ding and Ru Yang and Chang Guo and Guojian Zou and Bo Zhang and Meizi Li}
}

@misc{kashyap2025graphshypergraphsenhancingaspectbased,
    title={From Graphs to Hypergraphs: Enhancing Aspect-Based Sentiment Analysis via Multi-Level Relational Modeling}, 
    author={Omkar Mahesh Kashyap and Padegal Amit and Madhav Kashyap and Ashwini M Joshi and Shylaja SS},
    year={2025},
    eprint={2511.14142},
    archivePrefix={arXiv},
    primaryClass={cs.CL},
    url={https://arxiv.org/abs/2511.14142}, 
}

@inproceedings{NEURIPS2021_f1c15925,
    author = {Ying, Chengxuan and Cai, Tianle and Luo, Shengjie and Zheng, Shuxin and Ke, Guolin and He, Di and Shen, Yanming and Liu, Tie-Yan},
    booktitle = {Advances in Neural Information Processing Systems},
    editor = {M. Ranzato and A. Beygelzimer and Y. Dauphin and P.S. Liang and J. Wortman Vaughan},
    pages = {28877--28888},
    publisher = {Curran Associates, Inc.},
    title = {Do Transformers Really Perform Badly for Graph Representation?},
    url = {https://proceedings.neurips.cc/paper_files/paper/2021/file/f1c1592588411002af340cbaedd6fc33-Paper.pdf},
    volume = {34},
    year = {2021}
}

@misc{lei2025yolov13realtimeobjectdetection,
    title={YOLOv13: Real-Time Object Detection with Hypergraph-Enhanced Adaptive Visual Perception}, 
    author={Mengqi Lei and Siqi Li and Yihong Wu and Han Hu and You Zhou and Xinhu Zheng and Guiguang Ding and Shaoyi Du and Zongze Wu and Yue Gao},
    year={2025},
    eprint={2506.17733},
    archivePrefix={arXiv},
    primaryClass={cs.CV},
    url={https://arxiv.org/abs/2506.17733}, 
}

@article{ZHOU2020106292,
    title = {SK-GCN: Modeling Syntax and Knowledge via Graph Convolutional Network for aspect-level sentiment classification},
    journal = {Knowledge-Based Systems},
    volume = {205},
    pages = {106292},
    year = {2020},
    issn = {0950-7051},
    doi = {https://doi.org/10.1016/j.knosys.2020.106292},
    url = {https://www.sciencedirect.com/science/article/pii/S0950705120304688},
    author = {Jie Zhou and Jimmy Xiangji Huang and Qinmin Vivian Hu and Liang He}
}

@inproceedings{chen-etal-2020-inducing,
    title = "Inducing Target-Specific Latent Structures for Aspect Sentiment Classification",
    author = "Chen, Chenhua  and
      Teng, Zhiyang  and
      Zhang, Yue",
    editor = "Webber, Bonnie  and
      Cohn, Trevor  and
      He, Yulan  and
      Liu, Yang",
    booktitle = "Proceedings of the 2020 Conference on Empirical Methods in Natural Language Processing (EMNLP)",
    month = nov,
    year = "2020",
    address = "Online",
    publisher = "Association for Computational Linguistics",
    url = "https://aclanthology.org/2020.emnlp-main.451/",
    doi = "10.18653/v1/2020.emnlp-main.451",
    pages = "5596--5607"
}

@inproceedings{wang-etal-2020-relational,
    title = "Relational Graph Attention Network for Aspect-based Sentiment Analysis",
    author = "Wang, Kai  and
      Shen, Weizhou  and
      Yang, Yunyi  and
      Quan, Xiaojun  and
      Wang, Rui",
    editor = "Jurafsky, Dan  and
      Chai, Joyce  and
      Schluter, Natalie  and
      Tetreault, Joel",
    booktitle = "Proceedings of the 58th Annual Meeting of the Association for Computational Linguistics",
    month = jul,
    year = "2020",
    address = "Online",
    publisher = "Association for Computational Linguistics",
    url = "https://aclanthology.org/2020.acl-main.295/",
    doi = "10.18653/v1/2020.acl-main.295",
    pages = "3229--3238"
}

@inproceedings{tang-etal-2020-dependency,
    title = "Dependency Graph Enhanced Dual-transformer Structure for Aspect-based Sentiment Classification",
    author = "Tang, Hao  and
      Ji, Donghong  and
      Li, Chenliang  and
      Zhou, Qiji",
    editor = "Jurafsky, Dan  and
      Chai, Joyce  and
      Schluter, Natalie  and
      Tetreault, Joel",
    booktitle = "Proceedings of the 58th Annual Meeting of the Association for Computational Linguistics",
    month = jul,
    year = "2020",
    address = "Online",
    publisher = "Association for Computational Linguistics",
    url = "https://aclanthology.org/2020.acl-main.588/",
    doi = "10.18653/v1/2020.acl-main.588",
    pages = "6578--6588"
}

@inproceedings{li-etal-2021-dual-graph,
    title = "Dual Graph Convolutional Networks for Aspect-based Sentiment Analysis",
    author = "Li, Ruifan  and
      Chen, Hao  and
      Feng, Fangxiang  and
      Ma, Zhanyu  and
      Wang, Xiaojie  and
      Hovy, Eduard",
    editor = "Zong, Chengqing  and
      Xia, Fei  and
      Li, Wenjie  and
      Navigli, Roberto",
    booktitle = "Proceedings of the 59th Annual Meeting of the Association for Computational Linguistics and the 11th International Joint Conference on Natural Language Processing (Volume 1: Long Papers)",
    month = aug,
    year = "2021",
    address = "Online",
    publisher = "Association for Computational Linguistics",
    url = "https://aclanthology.org/2021.acl-long.494/",
    doi = "10.18653/v1/2021.acl-long.494",
    pages = "6319--6329"
}

@inproceedings{tian-etal-2021-aspect,
    title = "Aspect-based Sentiment Analysis with Type-aware Graph Convolutional Networks and Layer Ensemble",
    author = "Tian, Yuanhe  and
      Chen, Guimin  and
      Song, Yan",
    editor = "Toutanova, Kristina  and
      Rumshisky, Anna  and
      Zettlemoyer, Luke  and
      Hakkani-Tur, Dilek  and
      Beltagy, Iz  and
      Bethard, Steven  and
      Cotterell, Ryan  and
      Chakraborty, Tanmoy  and
      Zhou, Yichao",
    booktitle = "Proceedings of the 2021 Conference of the North American Chapter of the Association for Computational Linguistics: Human Language Technologies",
    month = jun,
    year = "2021",
    address = "Online",
    publisher = "Association for Computational Linguistics",
    url = "https://aclanthology.org/2021.naacl-main.231/",
    doi = "10.18653/v1/2021.naacl-main.231",
    pages = "2910--2922"
}

@inproceedings{zhang-etal-2022-ssegcn,
    title = "{SSEGCN}: Syntactic and Semantic Enhanced Graph Convolutional Network for Aspect-based Sentiment Analysis",
    author = "Zhang, Zheng  and
      Zhou, Zili  and
      Wang, Yanna",
    editor = "Carpuat, Marine  and
      de Marneffe, Marie-Catherine  and
      Meza Ruiz, Ivan Vladimir",
    booktitle = "Proceedings of the 2022 Conference of the North American Chapter of the Association for Computational Linguistics: Human Language Technologies",
    month = jul,
    year = "2022",
    address = "Seattle, United States",
    publisher = "Association for Computational Linguistics",
    url = "https://aclanthology.org/2022.naacl-main.362/",
    doi = "10.18653/v1/2022.naacl-main.362",
    pages = "4916--4925"
}

@inproceedings{chen-etal-2022-discrete,
    title = "Discrete Opinion Tree Induction for Aspect-based Sentiment Analysis",
    author = "Chen, Chenhua  and
      Teng, Zhiyang  and
      Wang, Zhongqing  and
      Zhang, Yue",
    editor = "Muresan, Smaranda  and
      Nakov, Preslav  and
      Villavicencio, Aline",
    booktitle = "Proceedings of the 60th Annual Meeting of the Association for Computational Linguistics (Volume 1: Long Papers)",
    month = may,
    year = "2022",
    address = "Dublin, Ireland",
    publisher = "Association for Computational Linguistics",
    url = "https://aclanthology.org/2022.acl-long.145/",
    doi = "10.18653/v1/2022.acl-long.145",
    pages = "2051--2064"
}

@article{Yu2023MWGCN,
    author  = {Yu, Bengong and Zhang, Shuwen},
    title   = {A novel weight-oriented graph convolutional network for aspect-based sentiment analysis},
    journal = {The Journal of Supercomputing},
    volume  = {79},
    number  = {1},
    pages   = {947--972},
    year    = {2023},
    doi     = {10.1007/s11227-022-04689-9},
    url     = {https://doi.org/10.1007/s11227-022-04689-9}
}

@inproceedings{ma-etal-2023-amr,
    title = "{AMR}-based Network for Aspect-based Sentiment Analysis",
    author = "Ma, Fukun  and
      Hu, Xuming  and
      Liu, Aiwei  and
      Yang, Yawen  and
      Li, Shuang  and
      Yu, Philip S.  and
      Wen, Lijie",
    editor = "Rogers, Anna  and
      Boyd-Graber, Jordan  and
      Okazaki, Naoaki",
    booktitle = "Proceedings of the 61st Annual Meeting of the Association for Computational Linguistics (Volume 1: Long Papers)",
    month = jul,
    year = "2023",
    address = "Toronto, Canada",
    publisher = "Association for Computational Linguistics",
    url = "https://aclanthology.org/2023.acl-long.19/",
    doi = "10.18653/v1/2023.acl-long.19",
    pages = "322--337"
}

@inproceedings{chen-etal-2024-s2gsl,
    title = "{S}$^2${GSL}: Incorporating Segment to Syntactic Enhanced Graph Structure Learning for Aspect-based Sentiment Analysis",
    author = "Chen, Bingfeng  and
      Ouyang, Qihan  and
      Luo, Yongqi  and
      Xu, Boyan  and
      Cai, Ruichu  and
      Hao, Zhifeng",
    editor = "Ku, Lun-Wei  and
      Martins, Andre  and
      Srikumar, Vivek",
    booktitle = "Proceedings of the 62nd Annual Meeting of the Association for Computational Linguistics (Volume 1: Long Papers)",
    month = aug,
    year = "2024",
    address = "Bangkok, Thailand",
    publisher = "Association for Computational Linguistics",
    url = "https://aclanthology.org/2024.acl-long.721/",
    doi = "10.18653/v1/2024.acl-long.721",
    pages = "13366--13379"
}

@article{JIN2025111654,
    title = {Aspect-based sentiment analysis with semantic and syntactic enhanced multi-layer fusion model},
    journal = {Engineering Applications of Artificial Intelligence},
    volume = {159},
    pages = {111654},
    year = {2025},
    issn = {0952-1976},
    doi = {https://doi.org/10.1016/j.engappai.2025.111654},
    url = {https://www.sciencedirect.com/science/article/pii/S0952197625016562},
    author = {Song Jin and Qing He and Yuji Wang and Nisuo Du and Wenjing Lei}
}

@inproceedings{10.1145/3331184.3331351,
    author = {Zhang, Chen and Li, Qiuchi and Song, Dawei},
    title = {Syntax-Aware Aspect-Level Sentiment Classification with Proximity-Weighted Convolution Network},
    year = {2019},
    isbn = {9781450361729},
    publisher = {Association for Computing Machinery},
    address = {New York, NY, USA},
    url = {https://doi.org/10.1145/3331184.3331351},
    doi = {10.1145/3331184.3331351},
    booktitle = {Proceedings of the 42nd International ACM SIGIR Conference on Research and Development in Information Retrieval},
    pages = {1145–1148},
    numpages = {4},
    location = {Paris, France},
    series = {SIGIR'19}
}

@inproceedings{dai-etal-2021-syntax,
    title = "Does syntax matter? A strong baseline for Aspect-based Sentiment Analysis with {R}o{BERT}a",
    author = "Dai, Junqi  and
      Yan, Hang  and
      Sun, Tianxiang  and
      Liu, Pengfei  and
      Qiu, Xipeng",
    editor = "Toutanova, Kristina  and
      Rumshisky, Anna  and
      Zettlemoyer, Luke  and
      Hakkani-Tur, Dilek  and
      Beltagy, Iz  and
      Bethard, Steven  and
      Cotterell, Ryan  and
      Chakraborty, Tanmoy  and
      Zhou, Yichao",
    booktitle = "Proceedings of the 2021 Conference of the North American Chapter of the Association for Computational Linguistics: Human Language Technologies",
    month = jun,
    year = "2021",
    address = "Online",
    publisher = "Association for Computational Linguistics",
    url = "https://aclanthology.org/2021.naacl-main.146/",
    doi = "10.18653/v1/2021.naacl-main.146",
    pages = "1816--1829"
}

@article{Feng2023PConv,
    author  = {Feng, Ao and Cai, Jiazhi and Gao, Zhengjie and Li, Xiaojie},
    title   = {Aspect-level sentiment classification with fused local and global context},
    journal = {Journal of Big Data},
    volume  = {10},
    number  = {1},
    pages   = {176},
    year    = {2023},
    doi     = {10.1186/s40537-023-00856-8},
    url     = {https://doi.org/10.1186/s40537-023-00856-8}
}

@inproceedings{wang-etal-2021-eliminating-sentiment,
    title = "Eliminating Sentiment Bias for Aspect-Level Sentiment Classification with Unsupervised Opinion Extraction",
    author = "Wang, Bo  and
      Shen, Tao  and
      Long, Guodong  and
      Zhou, Tianyi  and
      Chang, Yi",
    editor = "Moens, Marie-Francine  and
      Huang, Xuanjing  and
      Specia, Lucia  and
      Yih, Scott Wen-tau",
    booktitle = "Findings of the Association for Computational Linguistics: EMNLP 2021",
    month = nov,
    year = "2021",
    address = "Punta Cana, Dominican Republic",
    publisher = "Association for Computational Linguistics",
    url = "https://aclanthology.org/2021.findings-emnlp.258/",
    doi = "10.18653/v1/2021.findings-emnlp.258",
    pages = "3002--3012"
}

@article{JIAN2024103539,
    title = {Retrieval Contrastive Learning for Aspect-Level Sentiment Classification},
    journal = {Information Processing \& Management},
    volume = {61},
    number = {1},
    pages = {103539},
    year = {2024},
    issn = {0306-4573},
    doi = {https://doi.org/10.1016/j.ipm.2023.103539},
    url = {https://www.sciencedirect.com/science/article/pii/S0306457323002765},
    author = {Zhongquan Jian and Jiajian Li and Qingqiang Wu and Junfeng Yao}
}

@inproceedings{jian-etal-2025-agcl,
    title = "{AGCL}: Aspect Graph Construction and Learning for Aspect-level Sentiment Classification",
    author = "Jian, Zhongquan  and
      Wu, Daihang  and
      Wang, Shaopan  and
      Wang, Yancheng  and
      Yao, Junfeng  and
      Wang, Meihong  and
      Wu, Qingqiang",
    editor = "Rambow, Owen  and
      Wanner, Leo  and
      Apidianaki, Marianna  and
      Al-Khalifa, Hend  and
      Eugenio, Barbara Di  and
      Schockaert, Steven",
    booktitle = "Proceedings of the 31st International Conference on Computational Linguistics",
    month = jan,
    year = "2025",
    address = "Abu Dhabi, UAE",
    publisher = "Association for Computational Linguistics",
    url = "https://aclanthology.org/2025.coling-main.56/",
    pages = "841--854"
}

@inproceedings{li-etal-2021-learning-implicit,
    title = "Learning Implicit Sentiment in Aspect-based Sentiment Analysis with Supervised Contrastive Pre-Training",
    author = "Li, Zhengyan  and
      Zou, Yicheng  and
      Zhang, Chong  and
      Zhang, Qi  and
      Wei, Zhongyu",
    editor = "Moens, Marie-Francine  and
      Huang, Xuanjing  and
      Specia, Lucia  and
      Yih, Scott Wen-tau",
    booktitle = "Proceedings of the 2021 Conference on Empirical Methods in Natural Language Processing",
    month = nov,
    year = "2021",
    address = "Online and Punta Cana, Dominican Republic",
    publisher = "Association for Computational Linguistics",
    url = "https://aclanthology.org/2021.emnlp-main.22/",
    doi = "10.18653/v1/2021.emnlp-main.22",
    pages = "246--256"
}

@article{Ouyang_Yang_Liang_Wang_Wang_Li_2024, 
    title={Aspect-Based Sentiment Analysis with Explicit Sentiment Augmentations}, 
    volume={38}, 
    url={https://ojs.aaai.org/index.php/AAAI/article/view/29849}, 
    DOI={10.1609/aaai.v38i17.29849}, 
    abstractNote={Aspect-based sentiment analysis (ABSA), a fine-grained sentiment classification task, has received much attention recently. Many works investigate sentiment information through opinion words, such as \&amp;quot;good’’ and \&amp;quot;bad’’. However, implicit sentiment data widely exists in the ABSA dataset, whose sentiment polarity is hard to determine due to the lack of distinct opinion words. To deal with implicit sentiment, this paper proposes an ABSA method that integrates explicit sentiment augmentations (ABSA-ESA) to add more sentiment clues. We propose an ABSA-specific explicit sentiment generation method to create such augmentations. Specifically, we post-train T5 by rule-based data and employ three strategies to constrain the sentiment polarity and aspect term of the generated augmentations. We employ Syntax Distance Weighting and Unlikelihood Contrastive Regularization in the training procedure to guide the model to generate the explicit opinion words with the same polarity as the input sentence. Meanwhile, we utilize the Constrained Beam Search to ensure the augmentations are aspect-related. We test ABSA-ESA on two ABSA benchmarks. The results show that ABSA-ESA outperforms the SOTA baselines on implicit and explicit sentiment accuracy.}, 
    number={17}, 
    journal={Proceedings of the AAAI Conference on Artificial Intelligence}, 
    author={Ouyang, Jihong and Yang, Zhiyao and Liang, Silong and Wang, Bing and Wang, Yimeng and Li, Ximing}, 
    year={2024}, 
    month={Mar.}, 
    pages={18842–18850} 
}

@inproceedings{fei-etal-2023-reasoning,
    title = "Reasoning Implicit Sentiment with Chain-of-Thought Prompting",
    author = "Fei, Hao  and
      Li, Bobo  and
      Liu, Qian  and
      Bing, Lidong  and
      Li, Fei  and
      Chua, Tat-Seng",
    editor = "Rogers, Anna  and
      Boyd-Graber, Jordan  and
      Okazaki, Naoaki",
    booktitle = "Proceedings of the 61st Annual Meeting of the Association for Computational Linguistics (Volume 2: Short Papers)",
    month = jul,
    year = "2023",
    address = "Toronto, Canada",
    publisher = "Association for Computational Linguistics",
    url = "https://aclanthology.org/2023.acl-short.101/",
    doi = "10.18653/v1/2023.acl-short.101",
    pages = "1171--1182"
}

@article{Chang_Yang_Jiang_Xu_2024, 
    title={Counterfactual-Enhanced Information Bottleneck for Aspect-Based Sentiment Analysis}, 
    volume={38}, url={https://ojs.aaai.org/index.php/AAAI/article/view/29726}, 
    DOI={10.1609/aaai.v38i16.29726}, 
    abstractNote={Despite having achieved notable success for aspect-based sentiment analysis (ABSA), deep neural networks are susceptible to spurious correlations between input features and output labels, leading to poor robustness. In this paper, we propose a novel Counterfactual-Enhanced Information Bottleneck framework (called CEIB) to reduce spurious correlations for ABSA. CEIB extends the information bottleneck (IB) principle to a factual-counterfactual balancing setting by integrating augmented counterfactual data, with the goal of learning a robust ABSA model. Concretely, we first devise a multi-pattern prompting method, which utilizes the large language model (LLM) to generate high-quality counterfactual samples from the original samples. Then, we employ the information bottleneck principle and separate the mutual information into factual and counterfactual parts. In this way, we can learn effective and robust representations for the ABSA task by balancing the predictive information of these two parts. Extensive experiments on five benchmark ABSA datasets show that our CEIB approach achieves superior prediction performance and robustness over the state-of-the-art baselines. Code and data to reproduce the results in this paper is available at: https://github.com/shesshan/CEIB.}, 
    number={16}, 
    journal={Proceedings of the AAAI Conference on Artificial Intelligence}, 
    author={Chang, Mingshan and Yang, Min and Jiang, Qingshan and Xu, Ruifeng}, 
    year={2024}, 
    month={Mar.}, 
    pages={17736–17744} 
}

@inproceedings{yang-li-2024-modeling,
    title = "Modeling Aspect Sentiment Coherency via Local Sentiment Aggregation",
    author = "Yang, Heng  and
      Li, Ke",
    editor = "Graham, Yvette  and
      Purver, Matthew",
    booktitle = "Findings of the Association for Computational Linguistics: EACL 2024",
    month = mar,
    year = "2024",
    address = "St. Julian{'}s, Malta",
    publisher = "Association for Computational Linguistics",
    url = "https://aclanthology.org/2024.findings-eacl.13/",
    doi = "10.18653/v1/2024.findings-eacl.13",
    pages = "182--195"
}

@Article{app9163389,
    AUTHOR = {Zeng, Biqing and Yang, Heng and Xu, Ruyang and Zhou, Wu and Han, Xuli},
    TITLE = {LCF: A Local Context Focus Mechanism for Aspect-Based Sentiment Classification},
    JOURNAL = {Applied Sciences},
    VOLUME = {9},
    YEAR = {2019},
    NUMBER = {16},
    ARTICLE-NUMBER = {3389},
    URL = {https://www.mdpi.com/2076-3417/9/16/3389},
    ISSN = {2076-3417},
    DOI = {10.3390/app9163389}
}

@inproceedings{xing-etal-2020-tasty,
    title = "Tasty Burgers, Soggy Fries: Probing Aspect Robustness in Aspect-Based Sentiment Analysis",
    author = "Xing, Xiaoyu  and
      Jin, Zhijing  and
      Jin, Di  and
      Wang, Bingning  and
      Zhang, Qi  and
      Huang, Xuanjing",
    editor = "Webber, Bonnie  and
      Cohn, Trevor  and
      He, Yulan  and
      Liu, Yang",
    booktitle = "Proceedings of the 2020 Conference on Empirical Methods in Natural Language Processing (EMNLP)",
    month = nov,
    year = "2020",
    address = "Online",
    publisher = "Association for Computational Linguistics",
    url = "https://aclanthology.org/2020.emnlp-main.292/",
    doi = "10.18653/v1/2020.emnlp-main.292",
    pages = "3594--3605"
}

@InProceedings{10.1007/978-3-030-72240-1_7,
    author="Mukherjee, Rajdeep
    and Shetty, Shreyas
    and Chattopadhyay, Subrata
    and Maji, Subhadeep
    and Datta, Samik
    and Goyal, Pawan",
    editor="Hiemstra, Djoerd
    and Moens, Marie-Francine
    and Mothe, Josiane
    and Perego, Raffaele
    and Potthast, Martin
    and Sebastiani, Fabrizio",
    title="Reproducibility, Replicability and Beyond: Assessing Production Readiness of Aspect Based Sentiment Analysis in the Wild",
    booktitle="Advances in  Information Retrieval",
    year="2021",
    publisher="Springer International Publishing",
    address="Cham",
    pages="92--106",
    isbn="978-3-030-72240-1",
    doi = {10.1007/978-3-030-72240-1_7},
    url = {https://doi.org/10.1007/978-3-030-72240-1_7}
}

@inproceedings{10.1145/3583780.3614752,
    author = {Yang, Heng and Zhang, Chen and Li, Ke},
    title = {PyABSA: A Modularized Framework for Reproducible Aspect-based Sentiment Analysis},
    year = {2023},
    isbn = {9798400701245},
    publisher = {Association for Computing Machinery},
    address = {New York, NY, USA},
    url = {https://doi.org/10.1145/3583780.3614752},
    doi = {10.1145/3583780.3614752},
    booktitle = {Proceedings of the 32nd ACM International Conference on Information and Knowledge Management},
    pages = {5117–5122},
    numpages = {6},
    location = {Birmingham, United Kingdom},
    series = {CIKM '23}
}

@article{Feng_You_Zhang_Ji_Gao_2019, 
    title={Hypergraph Neural Networks}, 
    volume={33}, 
    url={https://ojs.aaai.org/index.php/AAAI/article/view/4235}, 
    DOI={10.1609/aaai.v33i01.33013558}, 
    abstractNote={\&amp;lt;p\&amp;gt;In this paper, we present a hypergraph neural networks (HGNN) framework for data representation learning, which can encode high-order data correlation in a hypergraph structure. Confronting the challenges of learning representation for complex data in real practice, we propose to incorporate such data structure in a hypergraph, which is more flexible on data modeling, especially when dealing with complex data. In this method, a hyperedge convolution operation is designed to handle the data correlation during representation learning. In this way, traditional hypergraph learning procedure can be conducted using hyperedge convolution operations efficiently. HGNN is able to learn the hidden layer representation considering the high-order data structure, which is a general framework considering the complex data correlations. We have conducted experiments on citation network classification and visual object recognition tasks and compared HGNN with graph convolutional networks and other traditional methods. Experimental results demonstrate that the proposed HGNN method outperforms recent state-of-theart methods. We can also reveal from the results that the proposed HGNN is superior when dealing with multi-modal data compared with existing methods.\&amp;lt;/p\&amp;gt;}, 
    number={01}, 
    journal={Proceedings of the AAAI Conference on Artificial Intelligence}, 
    author={Feng, Yifan and You, Haoxuan and Zhang, Zizhao and Ji, Rongrong and Gao, Yue}, 
    year={2019}, 
    month={Jul.}, 
    pages={3558–3565} 
}

@article{JI2026115040,
    title = {DAGF: A dual GCN and auxiliary graph fusion based model for aspect-based sentiment analysis},
    journal = {Applied Soft Computing},
    volume = {195},
    pages = {115040},
    year = {2026},
    issn = {1568-4946},
    doi = {https://doi.org/10.1016/j.asoc.2026.115040},
    url = {https://www.sciencedirect.com/science/article/pii/S1568494626004886},
    author = {Jie Ji and Wenlong Zhu and Chengle Hou and Qiaoyan Song and YuKun Ma and Youruo Wang}
}

@article{DBLP:journals/corr/abs-2006-03654,
  author = {Pengcheng He and
                  Xiaodong Liu and
                  Jianfeng Gao and
                  Weizhu Chen},
  title = {DeBERTa: Decoding-enhanced {BERT} with Disentangled Attention},
  journal = {CoRR},
  volume = {abs/2006.03654},
  year = {2020},
  url = {https://arxiv.org/abs/2006.03654},
  eprinttype = {arXiv},
  eprint = {2006.03654},
  bibsource = {dblp computer science bibliography, https://dblp.org}
}

@inproceedings{pontiki-etal-2014-semeval,
    title = "{S}em{E}val-2014 Task 4: Aspect Based Sentiment Analysis",
    author = "Pontiki, Maria  and
      Galanis, Dimitris  and
      Pavlopoulos, John  and
      Papageorgiou, Harris  and
      Androutsopoulos, Ion  and
      Manandhar, Suresh",
    editor = "Nakov, Preslav  and
      Zesch, Torsten",
    booktitle = "Proceedings of the 8th International Workshop on Semantic Evaluation ({S}em{E}val 2014)",
    month = aug,
    year = "2014",
    address = "Dublin, Ireland",
    publisher = "Association for Computational Linguistics",
    url = "https://aclanthology.org/S14-2004/",
    doi = "10.3115/v1/S14-2004",
    pages = "27--35"
}

@inproceedings{dong-etal-2014-adaptive,
    title = "Adaptive Recursive Neural Network for Target-dependent {T}witter Sentiment Classification",
    author = "Dong, Li  and
      Wei, Furu  and
      Tan, Chuanqi  and
      Tang, Duyu  and
      Zhou, Ming  and
      Xu, Ke",
    editor = "Toutanova, Kristina  and
      Wu, Hua",
    booktitle = "Proceedings of the 52nd Annual Meeting of the Association for Computational Linguistics (Volume 2: Short Papers)",
    month = jun,
    year = "2014",
    address = "Baltimore, Maryland",
    publisher = "Association for Computational Linguistics",
    url = "https://aclanthology.org/P14-2009/",
    doi = "10.3115/v1/P14-2009",
    pages = "49--54"
}

@inproceedings{pontiki-etal-2015-semeval,
    title = "{S}em{E}val-2015 Task 12: Aspect Based Sentiment Analysis",
    author = "Pontiki, Maria  and
      Galanis, Dimitris  and
      Papageorgiou, Haris  and
      Manandhar, Suresh  and
      Androutsopoulos, Ion",
    editor = "Nakov, Preslav  and
      Zesch, Torsten  and
      Cer, Daniel  and
      Jurgens, David",
    booktitle = "Proceedings of the 9th International Workshop on Semantic Evaluation ({S}em{E}val 2015)",
    month = jun,
    year = "2015",
    address = "Denver, Colorado",
    publisher = "Association for Computational Linguistics",
    url = "https://aclanthology.org/S15-2082/",
    doi = "10.18653/v1/S15-2082",
    pages = "486--495"
}

@inproceedings{pontiki-etal-2016-semeval,
    title = "{S}em{E}val-2016 Task 5: Aspect Based Sentiment Analysis",
    author = {Pontiki, Maria  and
      Galanis, Dimitris  and
      Papageorgiou, Haris  and
      Androutsopoulos, Ion  and
      Manandhar, Suresh  and
      AL-Smadi, Mohammad  and
      Al-Ayyoub, Mahmoud  and
      Zhao, Yanyan  and
      Qin, Bing  and
      De Clercq, Orph{\'e}e  and
      Hoste, V{\'e}ronique  and
      Apidianaki, Marianna  and
      Tannier, Xavier  and
      Loukachevitch, Natalia  and
      Kotelnikov, Evgeniy  and
      Bel, Nuria  and
      Jim{\'e}nez-Zafra, Salud Mar{\'i}a  and
      Eryi{\u{g}}it, G{\"u}l{\c{s}}en},
    editor = "Bethard, Steven  and
      Carpuat, Marine  and
      Cer, Daniel  and
      Jurgens, David  and
      Nakov, Preslav  and
      Zesch, Torsten",
    booktitle = "Proceedings of the 10th International Workshop on Semantic Evaluation ({S}em{E}val-2016)",
    month = jun,
    year = "2016",
    address = "San Diego, California",
    publisher = "Association for Computational Linguistics",
    url = "https://aclanthology.org/S16-1002/",
    doi = "10.18653/v1/S16-1002",
    pages = "19--30"
}

@inproceedings{jiang-etal-2019-challenge,
    title = "A Challenge Dataset and Effective Models for Aspect-Based Sentiment Analysis",
    author = "Jiang, Qingnan  and
      Chen, Lei  and
      Xu, Ruifeng  and
      Ao, Xiang  and
      Yang, Min",
    editor = "Inui, Kentaro  and
      Jiang, Jing  and
      Ng, Vincent  and
      Wan, Xiaojun",
    booktitle = "Proceedings of the 2019 Conference on Empirical Methods in Natural Language Processing and the 9th International Joint Conference on Natural Language Processing (EMNLP-IJCNLP)",
    month = nov,
    year = "2019",
    address = "Hong Kong, China",
    publisher = "Association for Computational Linguistics",
    url = "https://aclanthology.org/D19-1654/",
    doi = "10.18653/v1/D19-1654",
    pages = "6280--6285"
}

\appendix

\section{Experimental Details}
\label{appendix:experimental_details}

\subsection{Datasets}

We conduct experiments on six standard benchmark datasets on ABSA. The statistics of the six datasets are shown in Table~\ref{tab:main_datasets}. Restaurant14 and Laptop datasets are from SemEval-2014 Task 4 \citep{pontiki-etal-2014-semeval}. Twitter dataset is the target-dependent Twitter sentiment corpus introduced by \citet{dong-etal-2014-adaptive}. Restaurant15 and Restaurant16 datasets are from the restaurant-domain subsets of SemEval-2015 Task 12 and SemEval-2016 Task 5, respectively \citep{pontiki-etal-2015-semeval, pontiki-etal-2016-semeval}. MAMS is the Multi-Aspect Multi-Sentiment dataset introduced by \citet{jiang-etal-2019-challenge}, where each sentence contains multiple aspects with different sentiment polarities. All datasets are used for research purposes, following their
original licenses and distribution terms.

For ISE, we follow prior implicit-sentiment evaluation protocols and process all instances using the same dependency parsing and token-alignment pipelines as in the main experiments. For ARTS, we follow the adversarial aspect robustness evaluation protocol. Models are trained only on the original training split, without ARTS augmentation. The models are evaluated directly on the perturbed ARTS test samples.

\begin{table}[]
    \centering
    \resizebox{\columnwidth}{!}{
    \begin{tabular}{ccccc}
    \hline
    Dataset & Split & Positive & Neutral & Negative  \\
    \midrule
    \multirow{2}{*}{Restaurant14}
    & Train & 2164 & 637 & 807 \\
    & Test  & 727 & 196 & 196  \\
    \midrule
    \multirow{2}{*}{Laptop}
    & Train & 976 & 455 & 851 \\
    & Test  & 337 & 167 & 128 \\
    \midrule
    \multirow{2}{*}{Twitter}
    & Train & 1507 & 3016 & 1528 \\
    & Test  & 172  & 336  & 169  \\
    \midrule
    \multirow{2}{*}{Restaurant15}
    & Train & 912 & 36 & 256 \\
    & Test  & 326 & 34 & 182 \\
    \midrule
    \multirow{2}{*}{Restaurant16}
    & Train & 1240 & 69 & 439 \\
    & Test  & 469  & 30 & 117 \\
    \midrule
    \multirow{3}{*}{MAMS}
    & Train & 3380 & 5042 & 2764 \\
    & Val   & 403  & 604  & 325  \\
    & Test  & 400  & 607  & 329  \\
    \hline
    \end{tabular}
    }
    \caption{Statistics of the six standard benchmark datasets.}
    \label{tab:main_datasets}
\end{table}

\subsection{Implementation Details}

We use DeBERTa-v3-base as the pre-trained encoder and feed the sentence--aspect pair as encoder input. Word-level dependency parses are aligned to subword-level encoder states before graph construction. The model is optimized with AdamW using a batch size of 16 and gradient clipping of 1.0. The learning rate is selected from $\{1 \times 10^{-5}, 2 \times 10^{-5}, 2.5 \times 10^{-5}\}$, and the weight decay is selected from $\{1 \times 10^{-4}, 5 \times 10^{-3}\}$. We train for 30 epochs on six standard benchmark datasets.

For \textsc{GHI}, we use 2 reasoning layers. The static hyperedges include aspect, SRD-based local-context, and dependency hyperedges, with dependency hop threshold $\mathrm{T}=2$. The local window size is selected from $\{3,4\}$, the SRD radius from $\{3,5\}$, and the adaptive Top-$K$ from $\{3,4\}$. The number of adaptive hyperedges is set to $\mathrm{M}=6$ for Twitter and $\mathrm{M}=4$ for the rest. All experiments are conducted on a single NVIDIA RTX 5080 GPU.

\subsection{Additional Multi-seed Results on Twitter}

We additionally report multi-seed results on Twitter, as shown in Table~\ref{tab:twitter}. On the noisier Twitter domain, GHI obtains modest but consistent gains over the vanilla DeBERTa-base encoder, suggesting that the incidence reasoning layer is not limited to the cleaner SemEval-2014 domains.

\section{Framework Details}

\begin{table}[]
    \centering
    \resizebox{\columnwidth}{!}{
    \begin{tabular}{lcc}
    \hline
    \multirow{2}{*}{Model} 
    & \multicolumn{2}{c}{Twitter} \\

    \cmidrule(lr){2-3}
    
    & Acc & F1 \\
    \hline
    DeBERTa & 77.99(0.38) & 77.24(0.32) \\
    GHI(ours) & \textbf{78.31(0.38)} & \textbf{77.46(0.26)} \\
    \hline
    \end{tabular}
    }
    \caption{Additional controlled comparison against vanilla DeBERTa-base on Twitter. Results are reported as mean(std) over 5 seeds.}
    \label{tab:twitter}
\end{table}

\subsection{Adaptive Hyperedges Generation}

\begin{figure*}[t]
  \centering
  \includegraphics[width=0.85\linewidth]{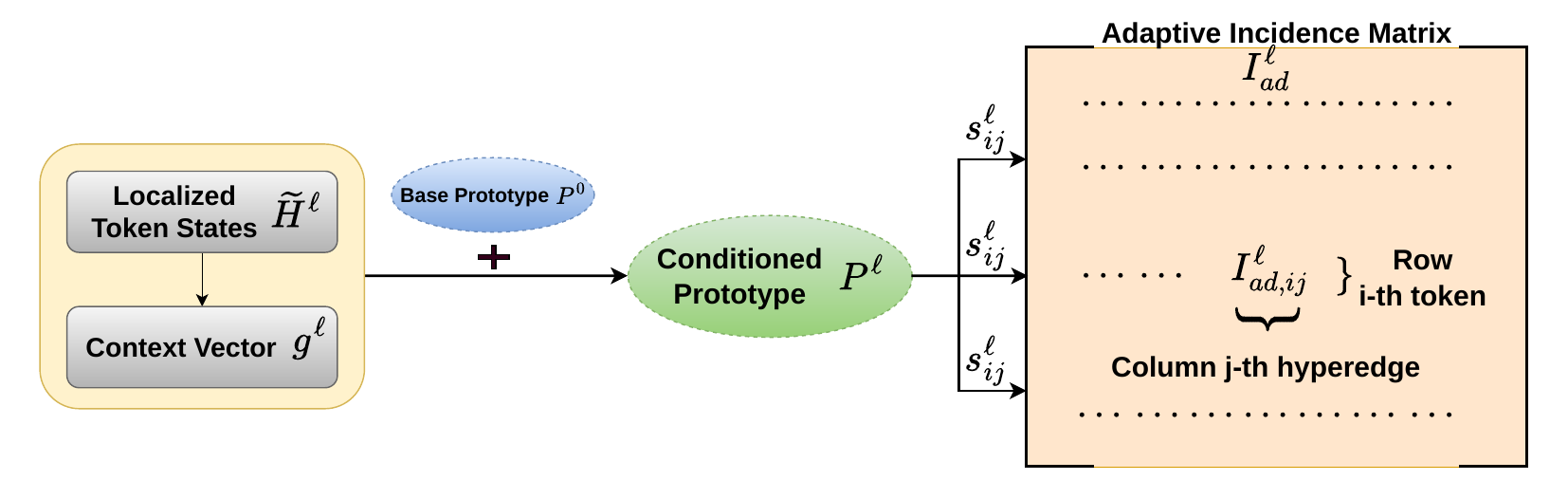}
  \caption{Adaptive Incidence Generation}
  \label{fig:adaptive_generation}
\end{figure*}

\label{appendix:adaptive_hyperedges_generating}
Let $P^0 \in \mathbb{R}^{M \times d_h}$ denote $M$ learnable base adaptive hyperedge prototypes, where $d_h$ denotes the hidden dimension. They are randomly initialized as global parameters and shared by all samples and reasoning layers, serving as the latent bases from which adaptive hyperedges are induced. 

At each reasoning layer, \textsc{GHI} conditions these shared prototypes on the current sentence and aspect anchors to induce sample-specific adaptive hyperedges. We first summarize the current instance by:

\begin{equation}
  \label{eq:9}
  g^{\ell} = [c^{\ell}; a^{\ell}; \mathrm{Pool}_{m}(\widetilde{H}^{\ell})],
\end{equation}

where $\mathrm{Pool}_m(\cdot)$ mean-pools valid sentence tokens. The context vector $g^\ell$ is mapped to a prototype offset $\Delta P^{\ell} \in \mathbb{R}^{M \times d}$, which is then used to generate the context-conditioned adaptive prototypes $P_j^\ell$:

\begin{equation}
  \label{eq:10}
  \Delta P^\ell = \mathrm{reshape}(\mathrm{MLP}(g^\ell)),
\end{equation}
\begin{equation}
  \label{eq:11}
  P_j^\ell =\mathrm{LN}(P_j^0 + \Delta P_j^\ell + a^\ell),\quad j=1,\ldots,M,
\end{equation}

where the aspect anchor is added to each prototype as a target-specific bias. Thus, sentence anchor $c^\ell$, aspect anchor $a^\ell$, and the pooled token context $\widetilde{H}^\ell$ adapt the shared bases $P^0$ to the current instance.

For each conditioned adaptive hyperedge prototype, GHI computes token--prototype participation scores for token $i\in V$ and
prototype $j=1,\ldots,M$:

\begin{equation}
  \label{eq:12}
  s_{ij}^{\ell} = \frac{\langle W_{t} \widetilde{H}_{i}^{\ell}, W_{p} P_{j}^{\ell} \rangle}{\sqrt{d_k}} + \frac{\langle W_{a} \widetilde{H}_{i}^{\ell}, a^{\ell} \rangle}{\sqrt{d_k}},
\end{equation}

where $W_t$, $W_p$, and $W_a$ are learned projections from tokens, prototypes and aspects, respectively. The $d_k$ denotes the projection dimension used for scaling.

The soft adaptive incidence is obtained by applying a masked softmax over valid graph-visible tokens for each prototype:

\begin{equation}
  \label{eq:13}
  I^{\ell}_{\mathrm{ad},ij} = \mathrm{MaskedSoftmax}_{i}(s^\ell_{ij};m),
\end{equation}

where $I^{\ell}_{ad,ij}$ is the normalized incidence weight between token $i$ and the adaptive hyperedge induced by prototype $j$, and the mask $m$ excludes invalid tokens from the normalization. Together with the prototype conditioning in Eqs.~\ref{eq:9}--\ref{eq:12}, this masked-softmax normalization instantiates $\mathrm{AdaptiveIncidence(\cdot)}$ in Eq.~\ref{eq:2}.

Finally, each conditioned prototype generates one adaptive hyperedge, and the $M$ generated hyperedges form $I^\ell_{\mathrm{ad}}\in\mathbb{R}^{|V|\times M}$. The overall generation process is illustrated in Figure~\ref{fig:adaptive_generation}.

\subsection{Layer-Wise Computation Flow}
\label{appendix:layer}

\begin{figure}[t]
  \includegraphics[width=\columnwidth]{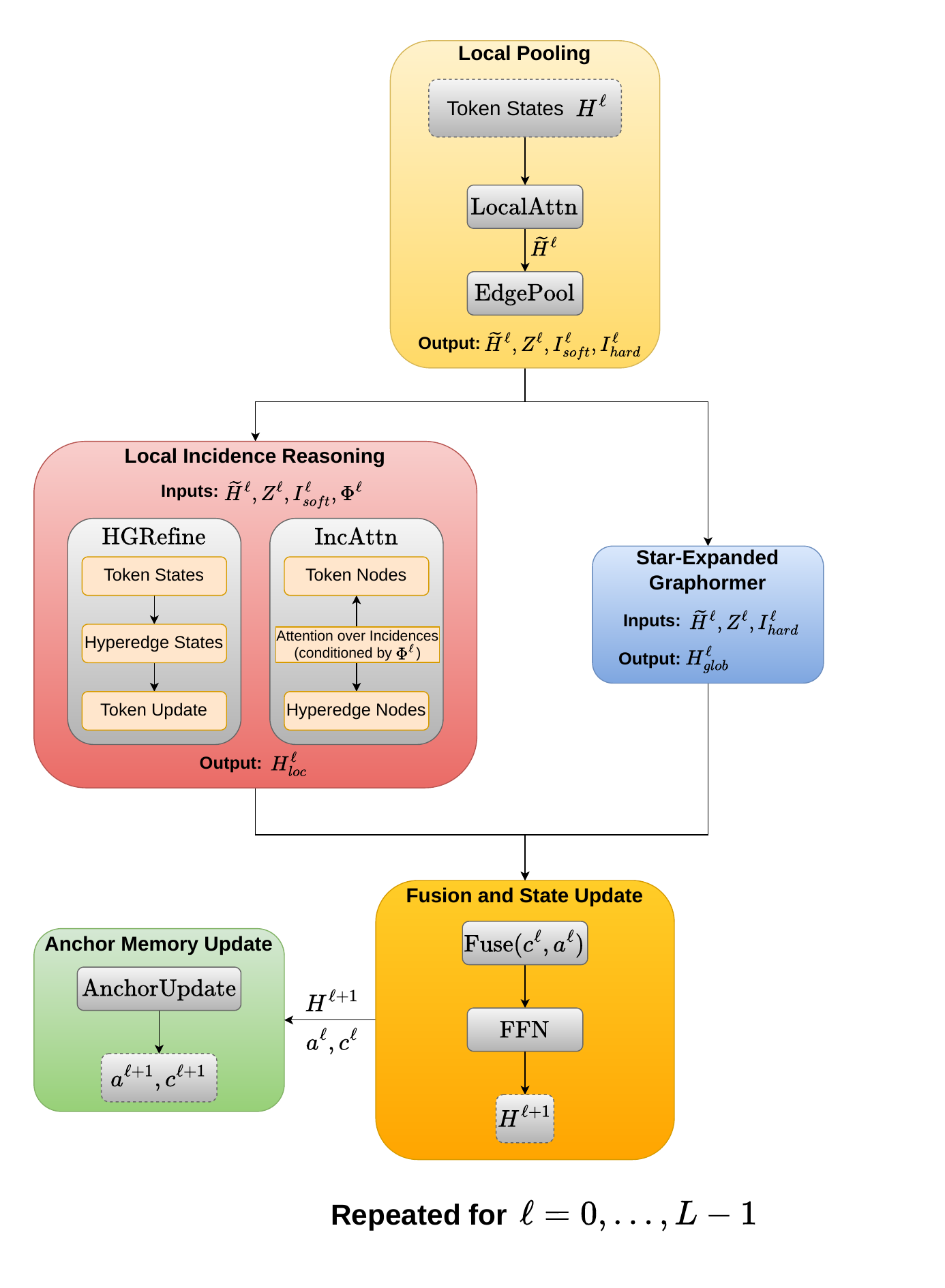}
  \caption{Layer-wise propagation in one GHI reasoning layer.}
  \label{fig:layer}
\end{figure}

As shown in Figure~\ref{fig:layer}, starting from token states $H^\ell$, local-window attention produces $\widetilde{H}^\ell$, edge pooling produces hyperedge states $Z^\ell$ and soft/hard incidence matrices. GHI then performs local incidence reasoning through $\mathrm{HGRefine}$ and $\mathrm{IncAttn}$ in parallel with star-expanded Graphormer reasoning, followed by anchor-conditioned fusion and a feed-forward update to obtain $H^{\ell+1}$. Anchor memories are updated from $H^{\ell+1}$ for the next layer. The details of the $\mathrm{HGRefine}$, $\mathrm{IncAttn}$ and $\mathrm{AnchorUpdate}$ are as follows.

\paragraph{HGRefine}

Given the soft incidence matrix $I^{\ell}_{soft}$, $\mathrm{HGRefine}$ performs incidence-weighted token--hyperedge propagation. We first column-normalize incidence weights over tokens for each hyperedge:

\begin{equation}
  \label{eq:14}
  \alpha^\ell_{ij} = \frac{I^\ell_{\mathrm{soft},ij}}{\sum_{i'} I^\ell_{\mathrm{soft},i'j}},
\end{equation}

and update hyperedge states by aggregating token states:

\begin{equation}
  \label{eq:15}
  \hat{Z}^\ell_j = \sum_i \alpha^\ell_{ij} W_t \widetilde{H}^\ell_i.
\end{equation}

The updated hyperedge states are then written back to tokens using row-normalized incidence weights $\beta^\ell_{ij}$ over hyperedges:

\begin{equation}
  \label{eq:16}
  \beta^\ell_{ij} = \frac{I^\ell_{\mathrm{soft},ij}}{\sum_{j'} I^\ell_{\mathrm{soft},ij'}},
\end{equation}
\begin{equation}
  \label{eq:17}
  v^\ell_i = W_o \sum_j \beta^\ell_{ij}\hat{Z}^\ell_j.
\end{equation}

The vector $v^{\ell}_i$ is the $\mathrm{HGRefine}$ message for token $i$ in Eq.~\ref{eq:4}.

\paragraph{IncAttn}

$\mathrm{IncAttn}$ further performs token--hyperedge attention with incidence-level relation features. For token $i$ and hyperedge $j$, we construct relation feature $\Phi^{\ell}_{ij}$ from edge type, edge role (static and adaptive), incidence weight, token SRD, and hyperedge SRD. For attention head $r$, $\mathrm{IncAttn}$ computes an attention logit $e^{\ell,r}_{ij}$ that measures how strongly token $i$ attends to hyperedge $j$:

\begin{equation}
  \label{eq:18}
  e^{\ell,r}_{ij} = \frac{\langle W_Q^r \widetilde{H}^\ell_i, W_K^r Z^\ell_j\rangle}{\sqrt{d_k}}+b_r(\Phi^\ell_{ij})+\log(I^\ell_{\mathrm{soft},ij}+\epsilon),
\end{equation}

where $W^r_Q$ and $W^r_K$ are the query and key projections for attention head $r$, $d_k$ is the projected key dimension, $b_r(\Phi^{\ell}_{ij})$ is the relation-aware attention bias, and $\epsilon$ is a small constant for numerical stability. The logarithmic term injects the soft incidence weight as a prior attention bias.

The attention weights are then obtained by masked softmax over valid hyperedges:

\begin{equation}
  \label{eq:19}
  a^{\ell,r}_{ij} = \mathrm{MaskedSoftmax}_{j}(e^{\ell,r}_{ij}).
\end{equation}

To modulate the value passed from each hyperedge, $\mathrm{IncAttn}$ further computes a scalar relation-aware gate:

\begin{equation}
  \label{eq:20}
  \gamma^\ell_{ij}=\sigma(W_\gamma \Phi^\ell_{ij}+b_\gamma),
\end{equation}

where $W_{\gamma}$ and $b_{\gamma}$ are learned parameters that map the relation feature $\Phi^{\ell}_{ij}$ to a scalar value gate. This gate controls how much the hyperedge $j$ contributes to the $\mathrm{IncAttn}$ message of token $i$.

The $\mathrm{IncAttn}$ message for token $i$ is then computed by aggregating gated hyperedge values across attention heads $r$:

\begin{equation}
  \label{eq:21}
  q^\ell_i = W_o \operatorname*{Concat}_{r} \sum_j a^{\ell,r}_{ij}\gamma^\ell_{ij} W_V^r Z^\ell_j.
\end{equation}

Here, $W^r_V$ is the value projection for head $r$, $W_o$ is the output projection, and $\operatorname*{Concat}_{r}$ concatenates the outputs of all heads $r$. The resulting $q^\ell_i$ is the $\mathrm{IncAttn}$ message added to the token update in Eq.~\ref{eq:4}.

\paragraph{AnchorUpdate}

After obtaining $H^{\ell+1}$, \textsc{GHI} updates the sentence and aspect anchors by gated residual MLPs. Let $\overline{H}^{\ell+1} = \mathrm{Pool}_m(H^{\ell+1})$ and $\overline{a}^{\ell+1} = \mathrm{Pool}_{[l,r)}(H^{\ell+1})$. We form:

\begin{equation}
  \label{eq:22}
  u_c^\ell = [c^\ell; a^\ell; \bar{h}^{\ell+1}],
\end{equation}
\begin{equation}
  \label{eq:23}
  u_a^\ell = [a^\ell; c^\ell; \bar{a}^{\ell+1}],
\end{equation}

and update the anchor as:

\begin{equation}
  \label{eq:24}
  c^{\ell+1} = c^\ell+\sigma(W_c u_c^\ell)\odot \mathrm{MLP}_c(u_c^\ell),
\end{equation}
\begin{equation}
  \label{eq:25}
  a^{\ell+1} = a^\ell+\sigma(W_a u_a^\ell)\odot \mathrm{MLP}_a(u_a^\ell).
\end{equation}

These updated anchors are used for next-layer incidence induction and local--global fusion.

\section{Sensitivity Analyses}
\label{appendix:sensitivity}

We further examine the sensitivity of \textsc{GHI} to three structural hyperparameters: the adaptive Top-$K$ used for constructing the sparse hard incidence topology, the number of adaptive hyperedges $M$, and the number of \textsc{GHI} reasoning layers $L$. We conduct the analysis on three representative datasets, Restaurant14, Laptop, and Twitter. For each study, we vary one hyperparameter from 1 to 6 while keeping the remaining settings fixed to the main configuration of the corresponding dataset.

\begin{figure*}[t]
  \includegraphics[width=\linewidth]{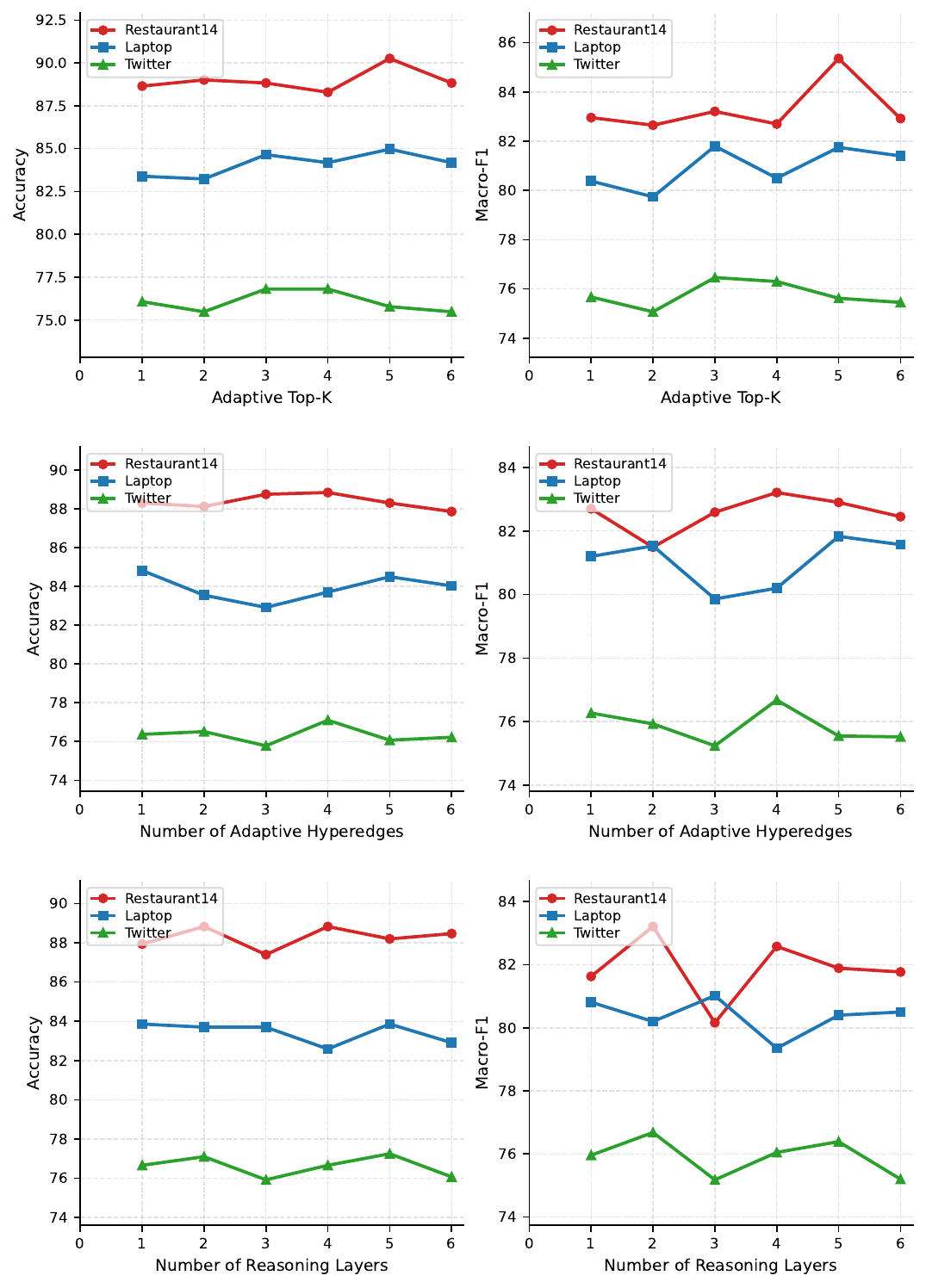}
  \caption{Sensitivity analyses for number of Adaptive Top-$K$, number of adaptive hyperedges $M$, and number of \textsc{GHI} layers $L$, respectively.}
  \label{fig:sensitivity}
\end{figure*}

Figure~\ref{fig:sensitivity} shows the results of the sensitivity experiments. For adaptive Top-$K$, moderate values generally perform better, suggesting that the hard topology benefits from retaining several high-order token representations. For the number of adaptive hyperedges $M$, performance usually saturates with a small number, indicating that a compact set of sample-specific evidence slots is sufficient for ABSA. Meanwhile, increasing $M$ further does not consistently improve performance and may introduce redundant or noisy memberships. For the number of reasoning layers $L$, \textsc{GHI} also performs well with shallow stacks, with $L=2$ or nearby values often yielding strong results. This supports our design choice of using compact incidence reasoning instead of relying on deep graph stacking.

\end{document}